\documentclass{article}
\usepackage{amsmath,epsfig,bm, amssymb,subfigure,graphicx,natbib}

\newcommand{\mathbbR}{\mathbb{R}}

\newcommand{\boldzero}{{\boldsymbol{0}}}
\newcommand{\boldone}{{\boldsymbol{1}}}

\newcommand{\boldA}{{\boldsymbol{A}}}

\newcommand{\boldD}{{\boldsymbol{D}}}

\newcommand{\boldI}{{\boldsymbol{I}}}

\newcommand{\boldK}{{\boldsymbol{K}}}
\newcommand{\boldL}{{\boldsymbol{L}}}
\newcommand{\boldM}{{\boldsymbol{M}}}

\newcommand{\boldV}{{\boldsymbol{V}}}
\newcommand{\boldW}{{\boldsymbol{W}}}
\newcommand{\boldX}{{\boldsymbol{X}}}

\newcommand{\boldd}{{\boldsymbol{d}}}

\newcommand{\boldm}{{\boldsymbol{m}}}

\newcommand{\boldu}{{\boldsymbol{u}}}
\newcommand{\boldv}{{\boldsymbol{v}}}
\newcommand{\boldw}{{\boldsymbol{w}}}
\newcommand{\boldx}{{\boldsymbol{x}}}
\newcommand{\boldy}{{\boldsymbol{y}}}

\newcommand{\boldalpha}{{\boldsymbol{\alpha}}}
\newcommand{\boldbeta}{{\boldsymbol{\beta}}}

\newcommand{\boldtheta}{{\boldsymbol{\theta}}}

\newcommand{\boldmu}{{\boldsymbol{\mu}}}

\newcommand{\boldphi}{{\boldsymbol{\phi}}}

\newcommand{\boldpsi}{{\boldsymbol{\psi}}}

\newcommand{\boldGamma}{{\boldsymbol{\Gamma}}}

\newcommand{\boldSigma}{{\boldsymbol{\Sigma}}}

\newcommand{\calX}{{\mathcal{X}}}
\newcommand{\calY}{{\mathcal{Y}}}

\newcommand{\pxy}{p_{\mathrm{x,y}}}

\usepackage{url,multirow}

\advance\oddsidemargin-1.0in
\date{\today}
\title{High-Dimensional Feature Selection by Feature-Wise Kernelized Lasso} 

\author{Makoto Yamada$^1$, Wittawat Jitkrittum$^2$, Leonid Sigal$^3$, Eric P. Xing$^4$, and Masashi Sugiyama$^{2}$\\
$^1$Yahoo! Labs, 701 1st Ave., Sunnyvale, CA, 94089, USA\\
$^2$Tokyo Institute of Technology, 2-12-1 O-okayama, Meguro-ku, Tokyo 152-8552, Japan\\
$^3$Disney Research Pittsburgh, 4720 Forbes Ave., Pittsburgh, PA 15213\\
$^4$Carnegie Mellon University, Pittsburgh, PA, 15213\\
\texttt{makotoy@yahoo-inc.com}, \texttt{wittawatj@gmail.com}, \\ \texttt{lsigal@disneyresearch.com}, \texttt{epxing@cs.cmu.edu}, \texttt{sugi@cs.titech.ac.jp}}

\begin{document}
\maketitle
\begin{abstract}
The goal of supervised feature selection is to find a
subset of input features that are responsible for predicting output values.
The \emph{least absolute shrinkage and selection operator} (Lasso)
allows computationally efficient feature selection based on linear dependency
between input features and output values.
In this paper, we consider a \emph{feature-wise} kernelized Lasso for capturing non-linear input-output dependency.
We first show that, with particular choices of kernel functions,
non-redundant features with strong statistical dependence on output values can be found in terms of kernel-based independence measures such as the Hilbert-Schmidt independence criterion (HSIC).
We then show that the globally optimal solution can be efficiently computed;
this makes the approach scalable to high-dimensional problems. The effectiveness of the proposed method is demonstrated through 
feature selection experiments for classification and regression with thousands of features.
\end{abstract}

\section{Introduction}
Finding a subset of features in high-dimensional supervised learning
is an important problem with many real-world applications such as gene selection from microarray data \citep{xing2001feature,JBCB:Ding+etal:2005,BMCBio:Suzuki+etal:2009a,huang2010variable}, document categorization \citep{CIKM:Forman:2008}, and prosthesis control \citep{IEEEBIO:Shenoy+etal:2008}. 

\subsection{Problem Description}
Let $\calX (\subset \mathbbR^d)$ be the domain of input vector $\boldx$
and $\calY (\subset \mathbbR)$ be the domain of output data\footnote{
$\calY$ could be either continuous (i.e., regression) or categorical (i.e., classification).
Structured outputs can also be handled in our proposed methods.
} $y$.
Suppose we are given $n$ independent and identically distributed (i.i.d.) paired samples,
\[
\{(\boldx_i, y_i)~|~\boldx_i \in \calX,~~y_i \in \calY,~i=1,\ldots,n\},
\]
drawn from a joint distribution with density $\pxy(\boldx, y)$. We denote the original data 
by 
\begin{align*}
\boldX &= [\boldx_1, \ldots, \boldx_n]\in \mathbbR^{d \times n},\\
\boldy &= [y_1, \ldots, y_n]^\top \in \mathbbR^{n},
\end{align*}
where $^\top$ denotes the transpose.

The goal of supervised feature selection is to find $m$ features ($m < d$) of input vector $\boldx$ that are responsible for predicting output $y$.

\subsection{Lasso}
The \emph{least absolute shrinkage and selection operator} (Lasso) \citep{JRSSB:Tibshirani:1996}
allows computationally efficient feature selection based on the assumption of linear dependency between input features and output values.

The Lasso optimization problem is given as 
\begin{align*}
\min_{\boldalpha \in \mathbbR^d} & \hspace{0.3cm} \frac{1}{2}\|\boldy - \boldX^\top \boldalpha \|^2_2 + \lambda \|\boldalpha\|_1,
\end{align*}
where $\boldalpha = [\alpha_1, \ldots, \alpha_d]^\top$ is a regression coefficient vector, $\alpha_k$ denotes the regression coefficient of the $k$-th feature, $\|\cdot\|_1$ and $\|\cdot\|_2$ are the $\ell_1$- and $\ell_2$-norms, and $\lambda > 0$ is the regularization parameter. The $\ell_1$-regularizer in Lasso tends to produce a sparse solution, which means that the regression coefficients for irrelevant features become zero. 
Lasso is particularly useful when the number of features is larger than the number of training samples \citep{JRSSB:Tibshirani:1996}.
Furthermore, various optimization software packages were developed for efficiently computing the Lasso solution
\citep{book:Boyd+Vandenberghe:2004,CommPAM:Daubechies+etal:2004,MMS:Combettes+Wajs:2005,IEEE-JSP:Kim+etal:2007,SIAM-JIS:Yin+etal:2008,IEEE-SP:Wright+etal:2009,JMLR:Tomioka+etal:2011}.

However, a critical limitation of Lasso is that it cannot capture  non-linear dependency.

\subsection{Instance-Wise Non-Linear Lasso}
To handle non-linearity, the \emph{instance-wise} non-linear Lasso was introduced \citep{IEEENN:Roth:2005}, where the original instance $\boldx$ is transformed by a non-linear function $\boldpsi(\cdot): \mathbbR^d \rightarrow \mathbbR^{d'}$.
Then the Lasso optimization problem is expressed as
\begin{align*}
\min_{\boldbeta \in \mathbbR^n} & \hspace{0.3cm} \frac{1}{2}\| \boldy  - \boldA \boldbeta \|^2_2 + \lambda \|\boldbeta\|_1,
\end{align*}
where $A_{i,j} = \boldpsi(\boldx_i)^\top \boldpsi(\boldx_j) = A(\boldx_i, \boldx_j)$,
$\boldbeta = [\beta_1, \ldots, \beta_n]^\top$ is a regression coefficient vector,
and $\beta_j$ is a coefficient of the $j$-th basis $A(\boldx,\boldx_j)$.

The instance-wise non-linear Lasso gives a sparse solution in terms of instances,
but not features.
Therefore, it cannot be used for feature selection.


\subsection{Feature-Wise Non-Linear Lasso (Feature Vector Machine)}
To obtain sparsity in terms of features,
the \emph{feature-wise} non-linear Lasso was proposed \citep{NIPS2005_644}.

The key idea is to apply a non-linear transformation in a feature-wise manner,
not in an instance-wise manner.
More specifically, let us represent the sample matrix $\boldX$ in a feature-wise manner
as
\[
\boldX= [\boldu_1,\ldots,\boldu_d]^\top\in \mathbbR^{d \times n},
\]
where $\boldu_k = [x_{k,1},\ldots,x_{k,n}]^\top \in \mathbbR^n$ is the vector of the $k$-th feature for all samples.
Then the feature vector $\boldu_k$
and the output vector $\boldy$ are transformed
by a non-linear function $\boldphi(\cdot): \mathbbR^n \rightarrow \mathbbR^{p}$.
The Lasso optimization problem in the transformed space is given as
\begin{align}
\min_{\boldalpha \in \mathbbR^d} & \hspace{0.3cm} \frac{1}{2}\|\boldphi(\boldy) - \sum_{k = 1}^d \alpha_k \boldphi(\boldu_k) \|^2_{2} + \lambda \|\boldalpha\|_1,
\label{eq:formulation-feature-wise-non-linear-Lasso}
\end{align}
where $\boldalpha = [\alpha_1, \ldots, \alpha_d]^\top$ is a regression coefficient vector
and $\alpha_k$ denotes the regression coefficient of the $k$-th feature.
By using the kernel trick \citep{book:Schoelkopf+Smola:2002},
Eq.\eqref{eq:formulation-feature-wise-non-linear-Lasso}
was shown to be equivalently expressed as the following quadratic programming (QP) problem:
\begin{align}
\min_{\boldalpha \in \mathbbR^d} &\hspace{0.3cm} \frac{1}{2} \boldalpha^\top \boldD \boldalpha, \nonumber \\
\textnormal{s.t.}& \hspace{0.3cm} \forall k, ~~| \boldalpha^\top  \boldd_k - D(\boldu_k,\boldy)| \leq \frac{\lambda}{2},
\label{eq:FVM-formulation} 
\end{align}
where
$D_{k,l} = \boldphi(\boldu_k)^\top \boldphi(\boldu_l) = D(\boldu_k, \boldu_l)$
and $\boldD = [\boldd_1,\ldots,\boldd_d]$.
This formulation is called the \emph{feature vector machine} (FVM). Note, since FVM uses $d \times d$ dimensional Hessian matrix $\boldD$, it is especially useful when the number of training samples $n$ is much bigger than that of features $d$.

In the original FVM, \emph{mutual information} \citep{book:Cover+Thomas:2006} 
was used as the kernel function $D(\boldu,\boldu')$.
However, the matrix $\boldD$ obtained from mutual information
is not necessarily positive definite \citep{NIPS:Seeger:2002},
and thus the objective function Eq.\eqref{eq:FVM-formulation} can be non-convex.
Furthermore, when the number of training samples is smaller than that of features
(which is often the case in high-dimensional feature selection scenarios),
the matrix $\boldD$ is singular.
This can cause numerical instability.
Another restriction of FVM is that,
irrespective of regression or classification,
output $\boldy$ should be transformed
by the same non-linear function $\boldphi(\cdot)$ as feature vector $\boldu$.
This highly limits the flexibility of capturing non-linear dependency.
Finally, it is not statistically clear what kind of features are found by this FVM formulation.

\subsection{Contribution of This Paper}
To overcome the limitations of FVM,
we propose an alternative feature-wise non-linear Lasso.
More specifically, we propose to use particular forms of 
\emph{universal reproducing kernels} \citep{JMLR:Steinwart:2001}
as feature and output transformations, and solve the optimization problem in the \emph{primal} space. 

An advantage of this new formulation is that
the global optimal solution can be computed efficiently.
Thus, it is scalable to high-dimensional feature selection problems. To the best of our knowledge, this is the first convex feature selection method that is able to deal with high-dimensional non-linearly related features. Furthermore, this new formulation  has a clear statistical interpretation that non-redundant features with strong statistical dependence on output values are found via kernel-based independence measures such as the \emph{Hilbert-Schmidt independence criterion} (HSIC) \citep{ALT:Gretton+etal:2005} and the criterion based on
the \emph{normalized cross-covariance operator} (NOCCO) \citep{NIPS:Fukumizu+etal:2008}. Thus, the proposed methods can be regarded as a \emph{minimum redundancy maximum relevance} based feature selection method \citep{PAMI:Peng+etal:2005}. In addition, the proposed methods are simple to implement, which is a highly preferable property for practitioners.

We also discuss the relation between the proposed method and existing feature selection approaches 
such as \emph{minimum redundancy maximum relevance} (mRMR) \citep{PAMI:Peng+etal:2005}, HSIC-based greedy feature selection \citep{song2012feature}, \emph{quadratic programming feature selection} (QPFS) \citep{JMLR:Rodriguez+etal:2010}, \emph{kernel target alignment} (KTA) \citep{DBLP:conf/nips/CristianiniSEK01,DBLP:journals/jmlr/CortesMR12}, Hilbert-Schmidt Feature Selection (HSFS) \citep{DBLP:conf/icml/MasaeliFD10}, and \emph{sparse additive models} (SpAM) \citep{ravikumar2009sparse, NIPS2008_0329, raskutti2012minimax}. See Table~\ref{tab:methods} for the summary of feature selection methods.

\begin{table*}[t]
  \centering
  \caption{Feature selection methods.}
  \label{tab:methods}
  \vspace*{2mm}
 \begin{tabular}{@{}c|ccccc@{}}
  \multirow{2}{*}{Method}&\multirow{2}{*}{Dependency}&\multirow{2}{*}{Optimization}           & \multirow{2}{*}{Primal/Dual} & Scalability  w.r.t.    &  \multirow{2}{*}{Structured output}                 
\\
   & & & & \# of features &  
\\
   \hline
   Lasso &Linear&{\bf Convex} & {\bf Primal}&{\bf Highly scalable} & Not available
\\
   mRMR &{\bf Non-linear}&Greedy & ---& Scalable & {\bf Available}
\\
   Greedy HSIC &{\bf Non-linear} &Greedy &--- & Scalable & {\bf Available} 
\\
  HSFS & {\bf Non-linear} & Non-convex & --- & Not scalable & {\bf Available} \\ 
   FVM  & {\bf Non-linear} &Non-convex$^\dagger$ & Dual &Not scalable & {\bf Available}
\\
   QPFS/KTA & {\bf Non-linear} &Non-convex$^\dagger$ & Dual &Not scalable & {\bf Available}
\\
   SpAM & Additive non-linear & {\bf Convex} & {\bf Primal} & Scalable & Not available
\\
    Proposed &{\bf Non-linear} & {\bf Convex} & {\bf Primal} &{\bf Highly scalable} & {\bf Available}
\\
\end{tabular}\\
\begin{flushleft}$^\dagger$In practice, positive constants may be added to the diagonal elements of
the Hessian matrix to guarantee the convexity, although the validity of selected features by this modification is not statistically clear. 
\end{flushleft}
\end{table*}

Through experiments on real-world feature selection problems,
we show that the proposed methods compare favorably with existing feature selection methods.

\section{Proposed Methods}
\label{sec:proposed}
In this section, we propose alternative implementations of
the non-linear feature-wise Lasso.


\subsection{HSIC Lasso}
We propose a feature-wise non-linear Lasso of the following form,
which we call the \emph{HSIC Lasso}\footnote{A MATLAB$^{\textregistered}$ implementation of the proposed algorithm is available from \url{http://www.makotoyamada-ml.com/hsiclasso.html}.}:
\begin{align}
\label{eq:HSIC Lasso-HSIC-relation}
\min_{\boldalpha \in \mathbbR^d} & \hspace{0.3cm}
\frac{1}{2}\|\bar{\boldL} - \sum_{k = 1}^{d} \alpha_k \bar{\boldK}^{(k)} \|^2_{\textnormal{Frob}}  +  \lambda \|\boldalpha\|_1, \nonumber \\
\textnormal{s.t.}&\hspace{0.3cm} \alpha_1,\ldots,\alpha_d \geq 0,
\end{align}
where $\|\cdot\|_{\textnormal{Frob}}$ is the Frobenius norm,
$\bar{\boldK}^{(k)} = \boldGamma \boldK^{(k)} \boldGamma$
and $\bar{\boldL} = \boldGamma \boldL \boldGamma$ are centered Gram matrices, 
$K^{(k)}_{i,j} = K(x_{k,i},x_{k,j})$ and $L_{i,j} = L(y_i,y_j)$ are Gram matrices,
$K(x,x')$ and $L(y,y')$ are kernel functions, 
$\boldGamma = \boldI_n - \frac{1}{n}\boldone_n \boldone_n^\top$ is the centering matrix,
$\boldI_n$ is the $n$-dimensional identity matrix, and $\boldone_n$ is the $n$-dimensional vector with all ones. 
Note that we employ non-negativity constraint for $\boldalpha$ so that meaningful features are selected (see Section~\ref{sec:statistical-interpretation} for details). In addition, since we use the output Gram matrix $\boldL$ to select features in HSIC Lasso, we can naturally incorporate structured outputs via kernels. Moreover, 
we can perform feature selection even if the training data set consists of input $\boldx$ and its affinity information $\boldL$ such as link structures between inputs. 

Differences from the original formulation \eqref{eq:formulation-feature-wise-non-linear-Lasso}
are that we allow the kernel functions $K$ and $L$ to be different
and the non-negativity constraint is imposed. 
The first term in Eq.\eqref{eq:HSIC Lasso-HSIC-relation} 
means that we are regressing the output kernel matrix $\bar{\boldL}$
by a linear combination of feature-wise input kernel matrices $\{\bar{\boldK}^{(k)}\}_{k=1}^d$. 

\subsection{Interpretation of HSIC Lasso}
Here, we show that HSIC Lasso can be regarded as a \emph{minimum redundancy maximum relevancy} (mRMR) based feature selection method  \citep{PAMI:Peng+etal:2005}, which is a popular feature selection strategy in machine learning and artificial intelligence communities.

\label{sec:statistical-interpretation}
The first term in Eq.\eqref{eq:HSIC Lasso-HSIC-relation} can be rewritten as
\begin{align}
\label{eq:HSICLasso_objective}
\frac{1}{2}\|\bar{\boldL} - \sum_{k = 1}^{d} \alpha_k \bar{\boldK}^{(k)} \|^2_{\textnormal{Frob}} & = \frac{1}{2} {\textnormal{HSIC}}(\boldy,\boldy) \!-\! \sum_{k = 1}^d\alpha_k {\textnormal{HSIC}}(\boldu_k,\boldy) \nonumber \\
&\phantom{=} \!+\! \frac{1}{2}\sum_{k,l = 1}^d \alpha_k \alpha_l {\textnormal{HSIC}}(\boldu_k,\boldu_l),
\end{align}
where ${\textnormal{HSIC}}(\boldu_k,\boldy) =\textnormal{tr}(\bar{\boldK}^{(k)}  \bar{\boldL} )$ is a kernel-based independence measure called the (empirical) \emph{Hilbert-Schmidt independence criterion} (HSIC) \citep{ALT:Gretton+etal:2005} and $\textnormal{tr}(\cdot)$ denotes the trace. ${\textnormal{HSIC}}(\boldy,\boldy)$ is a constant and can be ignored. HSIC always takes a non-negative value, and is zero if and only if two random variables are statistically independent when a \emph{universal reproducing kernel} \citep{JMLR:Steinwart:2001} such as the Gaussian kernel is used.
Note that the empirical HSIC asymptotically converges to the true HSIC with $O(1/\sqrt{n})$ (see Theorem~3 in \citet{ALT:Gretton+etal:2005}). In addition, HSIC can be regarded as the centered version of the \emph{kernel target alignment} (KTA) \citep{DBLP:conf/nips/CristianiniSEK01}.


If the $k$-th feature $\boldu_k$ has high dependence on output $\boldy$,
${\textnormal{HSIC}}(\boldu_k,\boldy)$ takes a large value and thus $\alpha_k$ should also take a large value so that Eq.\eqref{eq:HSIC Lasso-HSIC-relation} is minimized. On the other hand, if $\boldu_k$ is independent of $\boldy$, ${\textnormal{HSIC}}(\boldu_k,\boldy)$ is close to zero and thus such $\alpha_k$ tends to be eliminated by the $\ell_1$-regularizer. This means that relevant features that have strong dependence on output $\boldy$ tend to be selected by HSIC Lasso. 

Furthermore, if $\boldu_k$ and $\boldu_l$ are strongly dependent (i.e., redundant features),
${\textnormal{HSIC}}(\boldu_k,\boldu_l)$ takes a large value
and thus either of $\alpha_k$ and $\alpha_l$ tends to be zero.
This means that redundant features tend to be eliminated by HSIC Lasso.

Overall, HSIC Lasso tends to find non-redundant features with strong dependence
on output $\boldy$, which is the idea of \emph{minimum redundancy maximum relevancy} (mRMR) based feature selection methods \citep{PAMI:Peng+etal:2005}.
This is a preferable property in feature selection.

Note that, it is possible to remove the non-negativity constraint in Eq.\eqref{eq:HSIC Lasso-HSIC-relation} and select features that have non-zero coefficients $\boldalpha$. However, if we allow negative values in $\boldalpha$, it is hard to interpret selected features. Indeed, interpretability is one of important properties in feature selection, and thus we include the non-negativity constraint for HSIC Lasso. 

\subsection{Kernel Selection}

In theory, a universal kernel such as the Gaussian kernel or the Laplace kernel permits HSIC to detect dependence between two random variables \citep{ALT:Gretton+etal:2005}. Moreover, it has been proposed to use the delta kernel for multi-class classification problems \citep{song2012feature}. Thus, in this paper, we employ the Gaussian kernel for inputs. For output kernels, we use the Gaussian kernel for regression cases and the delta kernel for classification problems.

For input $x$, we first normalize the input $x$ to have unit standard deviation, and we use the Gaussian kernel: 
\[
K(x,x') = \exp \left(-\frac{( x - x' )^2}{2\sigma_{\mathrm x}^2} \right),
\]
where $\sigma_\mathrm{x}$ is the Gaussian kernel width. 

In regression scenarios (i.e., $y\in\mathbbR$), we normalize an output $y$ to have unit standard deviation, and we use the Gaussian kernel: 
\[
L(y,y') = \exp \left(-\frac{( y - y' )^2}{2\sigma_{\mathrm y}^2} \right),
\]
where  $\sigma_\mathrm{y}$ is the Gaussian kernel width. 

In classification scenarios (i.e., $y$ is categorical), we use the delta kernel for $y$, 
\begin{eqnarray*}
L(y,y') = \left\{ \begin{array}{ll}
{1}/{n_{y}} & \textnormal{if}~y = y', \\
0 & \textnormal{otherwise}, \\
\end{array} \right.
\end{eqnarray*} 
 where $n_{y}$ is the number of samples in class $y$. 
Note that it is also possible to use the Gaussian kernel in classification scenarios,
but it tends to perform poorly (see Figure \ref{fig:result_ASU_HSICLasso_outkernel}).

\subsection{Computational Properties of HSIC Lasso}
An important computational property of HSIC Lasso is that
the first term in Eq.\eqref{eq:HSIC Lasso-HSIC-relation} can be rewritten as
\begin{align*}
\frac{1}{2}\|\textnormal{vec}(\bar{\boldL}) - [\textnormal{vec}(\bar{\boldK}^{(1)}), \ldots,\textnormal{vec}(\bar{\boldK}^{(d)})] \boldalpha \|^2_{2},
\end{align*}
where $\textnormal{vec}(\cdot)$ is the vectorization operator.
This is the same form as plain Lasso
with $n^2$ samples and $d$ features.

If $d>n^2$ (i.e., high-dimensional feature selection from a small number of training samples),
the Lasso optimization technique called
the \emph{dual augmented Lagrangian} (DAL)\footnote{\url{http://www.ibis.t.u-tokyo.ac.jp/ryotat/dal/}} was shown to be computationally highly efficient
\citep{JMLR:Tomioka+etal:2011}.
Because DAL can also incorporate the non-negativity constraint
without losing its computational advantages,
we can directly use DAL to solve our HSIC Lasso problem. In contrast, when $n^2\ge d$, we may use the either cKTM or FVM (i.e., dual) formulation.


If the number of samples $n$ is relatively large, the Gaussian kernel computation in HSIC Lasso is expensive. Therefore, the overall computation cost of HSIC Lasso is high. In addition, since naive implementation of HSIC Lasso requires $n^2 d$ memory space, it is not practical if both $d$ and $n$ are large (e.g., $d > 10000$ and $n > 1000$). Here, we propose a table lookup approach to reduce the computation time and memory size.


More specifically, 
based on the fact that the Gaussian kernel values depend only on the difference of two input values,
we normalize every feature $x$ to have mean zero and unit standard deviation
and discretize the difference of two input values
into $B$ values (we use $B=4096$ in our implementation).
Then, we prepare in advance a lookup table of $B$ elements that contain Gaussian kernel values
and refer to these values when we compute the Gaussian kernels.
The centered Gram matrix $\bar{\boldK}$ can be rewritten as
\begin{align*}
\bar{\boldK} &= (\boldI - \frac{1}{n} \boldone \boldone^\top) \boldK (\boldI - \frac{1}{n} \boldone \boldone^\top) \\
&= \boldK - \boldone \boldm^\top - \boldm \boldone^\top + s \boldone \boldone^\top,
\end{align*}
where $\boldm = \frac{1}{n}\boldK \boldone \in \mathbbR^n$ and $s = \frac{1}{n^2}\boldone^\top \boldK \boldone \in \mathbbR$. Thus, to compute $\bar{\boldK}$, we only need to store $\boldm$ and $s$. Namely, the required memory size for the table lookup based approach is $O(dn + B)$, which is much smaller than the memory size required for the naive implementation, $O(dn^2)$.

 Another approach to deal with large sample size would be using stability selection which consists in running HSIC Lasso many times with subsampling and computing the number of times each feature is selected across the runs \citep{meinshausen2010stability,bach2008bolasso}.  


\subsection{Variation: NOCCO Lasso}
Instead of $\bar{\boldK}^{(k)}$ and $\bar{\boldL}$,
let us use 
$\widetilde{\boldK}^{(k)} = \bar{\boldK}^{(k)}(\bar{\boldK}^{(k)}+ \epsilon n \boldI_n)^{-1}$
and
$\widetilde{\boldL} = \bar{\boldL}(\bar{\boldL}+ \epsilon n \boldI_n)^{-1}$,
where $\epsilon > 0$ is a regularization parameter.
Then our optimization problem is expressed as
\begin{align*}
\min_{\boldalpha \in \mathbbR^d} & \hspace{0.3cm}
\frac{1}{2}\sum_{k,l = 1}^d \alpha_k \alpha_l{\textnormal{D}}^{\textnormal{NOCCO}}(\boldu_k,\boldu_l) - \sum_{k = 1}^d\alpha_k {\textnormal{D}}^{\textnormal{NOCCO}}(\boldu_k,\boldy) 
+\lambda \|\boldalpha\|_1, \\
\textnormal{s.t.}&\hspace{0.3cm} \alpha_1,\ldots,\alpha_d \geq 0,\nonumber
\end{align*}
where ${\textnormal{D}}^{\textnormal{NOCCO}}(\boldu_k,\boldy)  = \textnormal{tr}(\widetilde{\boldK}^{(k)} \widetilde{\boldL})$ is the kernel-based dependence measure based on
the (empirical) \emph{normalized cross-covariance operator} (NOCCO) \citep{NIPS:Fukumizu+etal:2008}.
We call this formulation the \emph{NOCCO Lasso}.

Because $\textnormal{D}_{\textnormal{NOCCO}}$ was shown to be asymptotically independent of
the choice of kernels, NOCCO Lasso is expected to be less sensitive to
the kernel parameter choice than HSIC Lasso,
although $\epsilon$ needs to be tuned in practice.
 
\subsection{Other Types of Regularizers}
The proposed method is amenable to most of the popular regularizers such as group-lasso and elastic-net regularizers \citep{meier2008group,zou2005regularization}, as well as to other feature selection problems. For example, the group-lasso regularizer can be easily incorporated into our framework as
\begin{align*}
\min_{\boldalpha \in \mathbbR^d} & \hspace{0.3cm}
\frac{1}{2}\|\bar{\boldL} - \sum_{k = 1}^{d} \alpha_k \bar{\boldK}^{(k)} \|^2_{\textnormal{Frob}}  +  \lambda \sum_{g = 1}^G \|\boldalpha_g\|_2, \nonumber \\
\textnormal{s.t.}&\hspace{0.3cm} \alpha_1,\ldots,\alpha_d \geq 0,
\end{align*}
where $\boldalpha = [\boldalpha_1^\top, \ldots, \boldalpha_G^\top]^\top$, $\boldalpha_g$ is the $g$th group of variables, and $G$ is the number of groups. This group-lasso problem can also be efficiently solved by the DAL package with the non-negativity constraint.


\subsection{Relation to Two-Stage Multiple Kernel Learning}
\label{sec:cKTA}
The proposed method is closely related to the two-stage \emph{multiple kernel learning} (MKL) method called centered kernel target alignment (cKTA) \citep{DBLP:journals/jmlr/CortesMR12}, which has been originally proposed for learning a kernel Gram matrix (not used for feature selection problems). 

 If we adopt cKTA for supervised feature selection problems, the optimization problem of cKTA can be written as
\begin{align*}
\min_{\boldalpha \in \mathbbR^d} &\hspace{0.3cm} \frac{1}{2}\sum_{k,l = 1}^d \alpha_k \alpha_l {\textnormal{HSIC}}(\boldu_k,\boldu_l) \!-\! \sum_{k = 1}^d\alpha_k {\textnormal{HSIC}}(\boldu_k,\boldy)  \nonumber \\
\textnormal{s.t.} &~~~\alpha_1,\ldots,\alpha_d \geq 0.
\end{align*}
Differences from the HSIC Lasso are that cKTA solve the optimization problem in the \emph{dual} space and does not have the $\ell_1$ regularization term.

An advantage of cKTA is that feature selection can be performed just by solving a non-negative least-squares problem. Moreover, since cKTA has $d\times d$ dimensional Hessian matrix ($\boldD_{kl} = {\textnormal{HSIC}}(\boldu_k,\boldu_l)$), cKTA is computationally efficient for feature selection problems with large $n$ and small $d$. However, the Hessian matrix $\boldD$ is not necessarily positive definite \citep{NIPS:Seeger:2002} and is singular in high-dimensional problems.
More specifically, $\boldD$ can be written as
\[
\boldD \!\!=\!\!  [\textnormal{vec}(\bar{\boldK}^{(1)}\!), \!\ldots\! ,\textnormal{vec}(\bar{\boldK}^{(d)}\!)]^\top [\textnormal{vec}(\bar{\boldK}^{(1)}\!), \!\ldots\! ,\textnormal{vec}(\bar{\boldK}^{(d)}\!)],
\]
and $\boldD$ is singular when $d>n^2$ (i.e., high-dimensional feature selection from a small number of training samples). Thus, solving the non-negative least-squares problem in high-dimensional feature selection problems can be cumbersome in practice. 


For the above reason, the proposed feature selection method is more suited than cKTA for high-dimensional feature selection problems. In contrast, if we want to solve a small $d$ and large $n$ feature selection problem, cKTA is more suited than the HSIC Lasso.

\section{Existing Methods}
\label{sec:existing}
In this section, we review existing feature selection methods
and discuss their relation to the proposed approach.
See Table~\ref{tab:methods} for the summary of feature selection methods.

\subsection{Minimum Redundancy Maximum Relevance (mRMR)}
\emph{Minimum redundancy maximum relevance} (mRMR) \citep{PAMI:Peng+etal:2005}
is a mutual information based feature selection criterion. 


Let $\boldV=[\boldv_1,\ldots,\boldv_m]^\top \in \mathbbR^{m \times n}$ be a sub-matrix of $\boldX= [\boldu_1,\ldots,\boldu_d]^\top\in \mathbbR^{d \times n}$,
where $m$ features are extracted from $d$ features.
Then mRMR for $\boldV$ is defined as follows:
\begin{align}
\textnormal{mRMR}(\boldV) &= 
\frac{1}{m}\sum_{k = 1}^m \widehat{\textnormal{MI}}(\boldv_k,\boldy)
 - \frac{1}{m^2}\sum_{k,l = 1}^m\widehat{\textnormal{MI}}(\boldv_k, \boldv_l),
\label{eq:mRMR}
\end{align}
where $\widehat{\textnormal{MI}}(\boldv,\boldy)$ is an empirical 
approximator of mutual information given as
\begin{align*}
\widehat{\textnormal{MI}}(\boldv,\boldy)
= \iint \widehat{p}_{\mathrm{v,y}}(\boldv,\boldy) \log \frac{\widehat{p}_{\mathrm{v,y}}(\boldv,\boldy)}{\widehat{p}_{\mathrm{v}}(\boldv)\widehat{p}_{\mathrm{y}}(\boldy)} \textnormal{d}\boldv \textnormal{d}\boldy.
\end{align*}
$\widehat{p}_{\mathrm{v,y,}}(\boldv,\boldy)$ denotes a Parzen window estimator of
the joint density of $\boldv$ and $\boldy$, and 
$\widehat{p}_{\mathrm{v}}(\boldv)$ and $\widehat{p}_{\mathrm{y}}(\boldy)$ denotes
Parzen window estimators of marginal densities of $\boldv$ and $\boldy$, respectively.

The first term in mRMR measures the dependency between chosen feature $\boldv_k$ and output $\boldy$,
while the second term is a penalty for selecting redundant features. Thus, mRMR finds non-redundant features with strong dependence on outputs. A very fast implementation of mRMR available, and thus, it can deal with high-dimensional feature selection problems.

mRMR-based feature selection is performed by
finding a sub-matrix $\boldV$ that maximizes Eq.\eqref{eq:mRMR}.
However, since there are $2^d$ possible feature subsets,
the brute force approach is computationally intractable.
Hence, greedy search strategies such as forward selection/backward elimination
are used in practice \citep{PAMI:Peng+etal:2005}. 
However, the greedy approaches tend to produce a locally optimal feature set.

Another potential weakness of mRMR is that mutual information is approximated
by Parzen window estimation---Parzen window based mutual information estimation is unreliable
when the number of training samples is small \citep{BMCBio:Suzuki+etal:2009a}.

\subsection{Greedy Feature Selection with HSIC}
\label{sec:application-mutualinformation-independence-HSIC}

The HSIC-based feature selection criterion \citep{song2012feature} is defined as
\begin{align}
\textnormal{tr}(\bar{\boldM} \bar{\boldL}),
\label{eq:HSFS}
\end{align}
where
$\bar{\boldM} = \boldGamma \boldM \boldGamma$ is a centered Gram matrix, 
$M_{i,j} = M(\boldv_{i},\boldv_{j})$ is a Gram matrix, 
$M(\boldv,\boldv')$ is a kernel function,
and
$(\boldv_1,\ldots,\boldv_m)=\boldV\in \mathbbR^{m \times n}$.

HSIC-based greedy feature selection is performed 
by finding a sub-matrix $\boldV$ that maximizes Eq.\eqref{eq:HSFS}.
An advantage of HSIC-based feature selection is its simplicity;
it can be implemented very easily.
However, since the maximization problem Eq.\eqref{eq:HSFS} is NP-hard,
forward selection/backward elimination strategies are used for finding a locally optimal solution
in practice \citep{song2012feature}.

\subsection{Hilbert-Schmidt Feature Selection (HSFS)}
\emph{Hilbert-Schmidt feature selection} (HSFS) \citep{DBLP:conf/icml/MasaeliFD10} is defined as
\begin{align*}
\min_{\boldW \in \mathbbR^{d\times d}} &\hspace{0.3cm} -\textnormal{HSIC}(\boldW \boldX, \boldy) + \lambda \sum_{j = 1}^d\|\boldw_j\|_{\infty},
\end{align*}
where $\boldW = [\boldw_1, \ldots, \boldw_d]$ is a transformation matrix, $\lambda > 0$ is the regularization parameter, and $\|\cdot\|_{\infty}$ is the $\ell_{\infty}$-norm. 

HSFS can be regarded as a continuously relaxed version of the HSIC-based feature selection \citep{song2012feature}.
Thanks to this continuous formulation,
the HSFS optimization problem can be solved by limited-memory BFGS (L-BFGS)
\citep{book:Nocedal:2003}.  
However, since HSFS is a non-convex method, restarting from many different initial points would be necessary to select good features, which is computationally expensive. Moreover, HSFS attempts to optimize a projection matrix which has $d^2$ parameters. That is, following the original HSFS implementation based on a Quasi-Newton method, the total computational complexity of HSFS is $O(d^4)$, which can be unacceptably large in high-dimensional feature selection problems. To reduce the computational cost, it may be able to approximately solve the HSFS optimization problem by reducing the size of the transformation matrix to $d\times q$ for $q \ll d$. However, this approximation leads to an additional tuning parameter that can not be chosen objectively.

\subsection{Quadratic Programming Feature Selection (QPFS)}
\label{sec:QPFS}
\emph{Quadratic programming feature selection} (QPFS) \citep{JMLR:Rodriguez+etal:2010} 
tries to find features by solving a QP problem.

The QPFS optimization problem is defined as
\begin{align*}
\min_{\boldtheta \in \mathbbR^d} &\hspace{0.3cm} \frac{(1-\gamma)}{2} \boldtheta^\top \boldD \boldtheta - \gamma \boldtheta^\top \boldd, \nonumber \\
\textnormal{s.t.} &~~~ \sum_{k = 1}^d \theta_k = 1, ~~~\theta_1,\ldots,\theta_d \geq 0,
\end{align*}
where $\boldtheta = [\theta_1, \ldots, \theta_d]^\top$, $D_{k,l} = D(\boldu_k,\boldu_l)$, $\boldd = [D(\boldu_1,\boldy),\ldots, D(\boldu_d,\boldy)]^\top$, $D(\cdot)$ is a dependency measure, and $\gamma \in [0,1]$ is a tuning parameter. In QPFS, 
an empirical estimator of mutual information is used as a dependency measure. Note that if we employ HSIC as a dependency measure in QPFS and remove the sum-to-one constraint, QPFS is equivalent to the centered KTA (cKTA) \citep{DBLP:journals/jmlr/CortesMR12}, which is a multiple kernel learning method and has been originally proposed for learning a kernel matrix (see Section \ref{sec:cKTA} for details).

Similar to cKTA, an advantage of QPFS is that feature selection can be performed just by solving a QP problem. Moreover, since QPFS has $d\times d$ dimensional Hessian matrix, QPFS is computationally efficient
for feature selection problems with large $n$ and small $d$. However, 
the Hessian matrix $\boldD$ is not necessarily positive definite \citep{NIPS:Seeger:2002}
and is singular in high-dimensional problems.

\subsection{Sparse Additive Models (SpAM)}
The sparse additive models (SpAM)  is useful for high-dimensional feature selection \citep{ravikumar2009sparse,NIPS2008_0329, raskutti2012minimax,suzuki2012fast}.

The SpAM optimization problem can be expressed as
\begin{align*}
\min_{\boldbeta_1,\ldots, \boldbeta_d \in \mathbbR^n} & \hspace{0.3cm} \| \boldy  \!-\! \sum_{k = 1}^d \boldK^{(k)} \boldbeta_k \|^2_2 + \lambda  \sum_{k = 1}^d \!\sqrt{\frac{1}{n} \| \boldK^{(k)}\boldbeta_k\|^2_2},
\end{align*}
where 
 $\boldbeta_k = [\beta_{k,1}, \ldots, \beta_{k,n}]^\top, k = 1,\ldots,d$ are regression coefficient vectors, $\beta_{k,j}$ is a coefficient for $[K(x_{k,1},x_{k,j}),\ldots,K(x_{k,n},x_{k,j})]^\top$\sloppy, and $\lambda > 0$ is a regularization parameter. This problem can be efficiently solved by the \emph{back-fitting} algorithm \citep{ravikumar2009sparse}. Note that SpAM is closely related to the \emph{hierarchical multiple kernel learning} \citep{NIPS2008_0171}, which employs a sparse additive model with an alternative sparsity-inducing regularizer. 

An advantage of SpAM is that it is a convex method and can be efficiently optimized by the backfitting algorithm. Moreover, statistical properties of the SpAM estimator are well studied \citep{ravikumar2009sparse}. A potential weakness of SpAM is that it can only deal with additive models. That is, if data follows a non-additive model, SpAM may not work well (see Figure~\ref{fig:illustrative_example}(b)). 
Another weakness of SpAM is that it needs to optimize $nd$ variables,
while the proposed methods contain only $d$ variables.
Thus, SpAM optimization tends to be computationally more expensive than the proposed methods.
 Finally, an output $y$ should be a real number in SpAM, meaning that SpAM cannot deal with structured outputs such as multi-label and graph data. 


\section{Experiments}
\label{sec:experiments}
In this section, we experimentally investigate the performance of the proposed and existing feature selection methods using synthetic and real-world datasets. 

\subsection{Setup}


We compare the performance of the proposed methods with mRMR \citep{PAMI:Peng+etal:2005}, QPFS \citep{JMLR:Rodriguez+etal:2010}, cKTA \citep{DBLP:journals/jmlr/CortesMR12}, forward selection with HSIC (FHSIC), FVM \citep{NIPS2005_644}, and SpAM\footnote{We thank the authors of \citep{ravikumar2009sparse} for providing us the code used in their paper.} \citep{ravikumar2009sparse}. Note that, since it has been reported that the performance of FHSIC is comparable to HSFS and HSFS is computationally expensive for high-dimensional data, we decided to only compare the proposed method to FHSIC. For FVM, QPFS, and mRMR, the C++ implementation of a mutual information estimator\footnote{\url{http://penglab.janelia.org/proj/mRMR/}} is used. Then, a QP solver SeDuMi\footnote{\url{http://sedumi.ie.lehigh.edu/}} is used to solve the FVM and QPFS optimization problems. 
We observed that the matrices $\boldD$ in FVM, QPFS, and cKTA tend not to be positive definite.
In our experiments, we added a small constant to the diagonal elements of $\boldD$ so that the objective function becomes convex. For all experiments, we set $\lambda = 1$ in FVM, $\gamma = 0.5$ in QPFS and cKTA, and $\epsilon = 10^{-3}$ in NOCCO Lasso. For proposed methods, we experimentally use $\sigma_{\mathrm x} = 1$.

\subsection{Synthetic Datasets}
First, we illustrate the behavior of the proposed HSIC Lasso and NOCCO Lasso
using the following two synthetic datasets:
\begin{description}
\item[(a) Data1 (Additive model):] 
\[
Y = -2\sin(2X_1) + X_2^2 + X_3 + \exp(-X_4) + E,
\] 
where $(X_1, \ldots, X_{256})^\top \sim N(\boldzero_{256},\boldI_{256})$ and $E \sim N(0,1)$. Here, $N(\boldmu,\boldSigma)$ denotes the multi-variate Gaussian distribution with mean $\boldmu$ and covariance matrix $\boldSigma$.
\item[(b) Data2 (Non-additive model):] 
\[
Y = X_{1}\exp(2X_{2}) + X_3^2 + E,
\] 
where $(X_1, \ldots, X_{1000})^\top \sim N(\boldzero_{1000},\boldI_{1000})$ and $E \sim N(0,1)$.
\end{description} 

\begin{figure*}[t!]
\begin{center}
\begin{minipage}[t]{0.45\linewidth}
\centering
  {\includegraphics[width=0.99\textwidth]{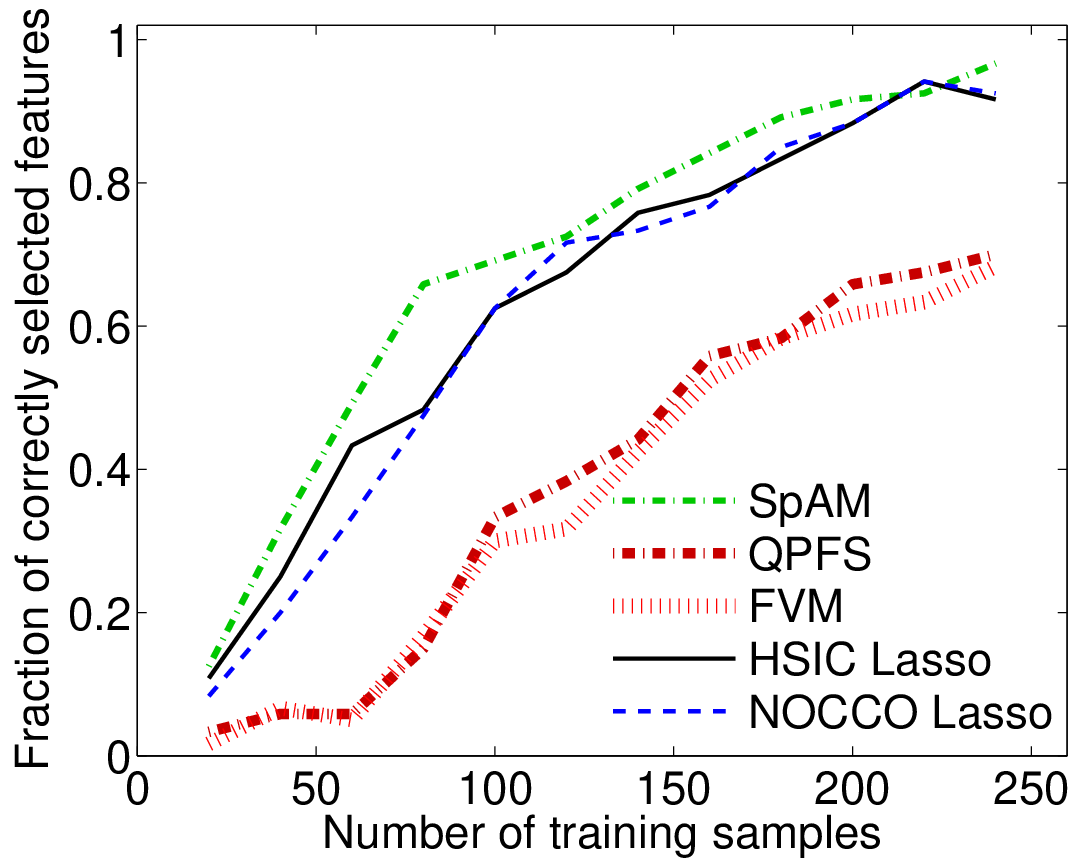}} \\ \vspace{-0.10cm}
(a) Data1 (Additive model)
\end{minipage}
\begin{minipage}[t]{0.45\linewidth}
\centering
  {\includegraphics[width=0.99\textwidth]{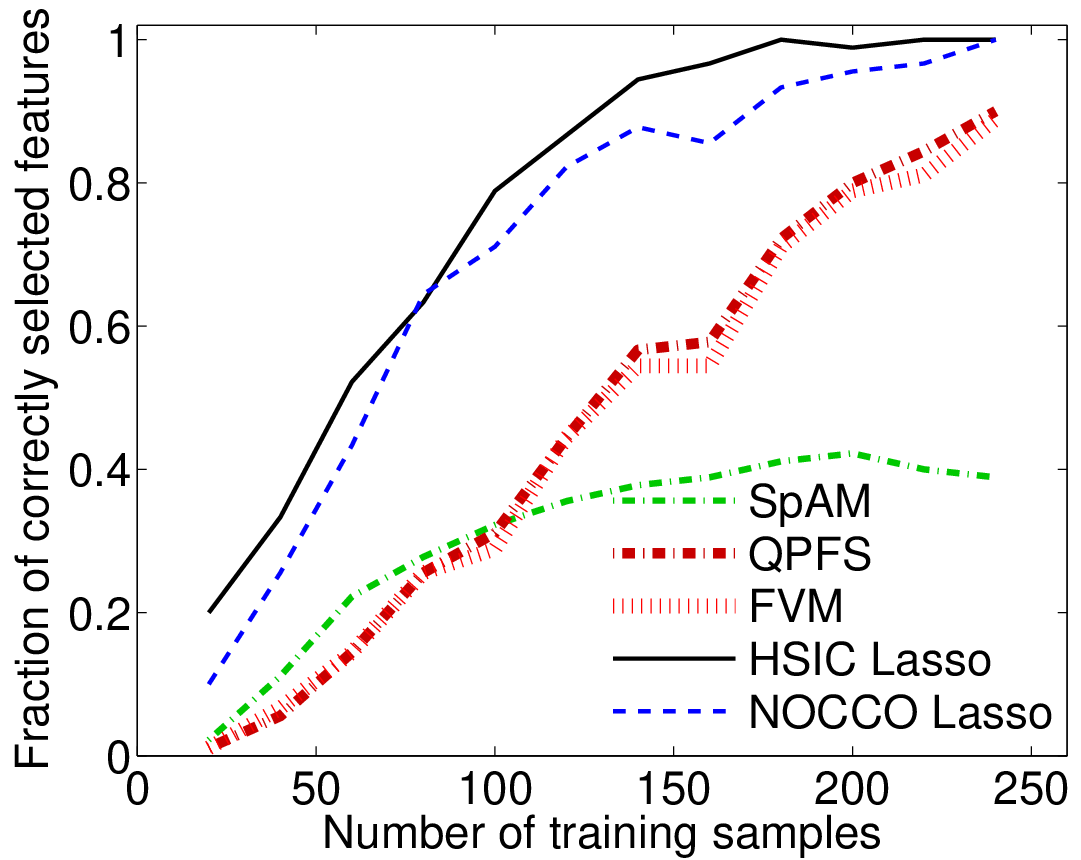}} \\ \vspace{-0.10cm}
(b) Data2 (Non-additive model)
\end{minipage}\\
\begin{minipage}[t]{0.475\linewidth}
\centering
  {\includegraphics[width=0.99\textwidth]{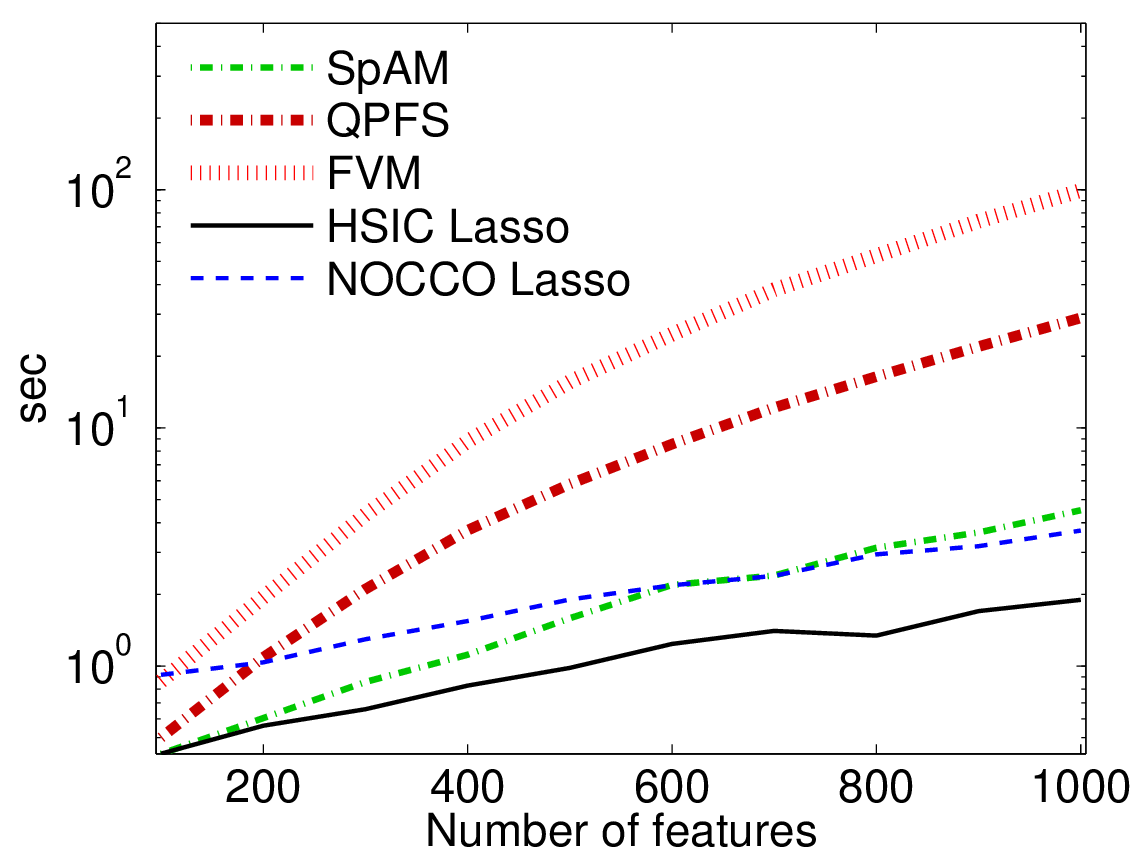}} \\ \vspace{-0.10cm}
(c) Computation time for Data2
\end{minipage}
\begin{minipage}[t]{0.475\linewidth}
\centering
  {\includegraphics[width=0.99\textwidth]{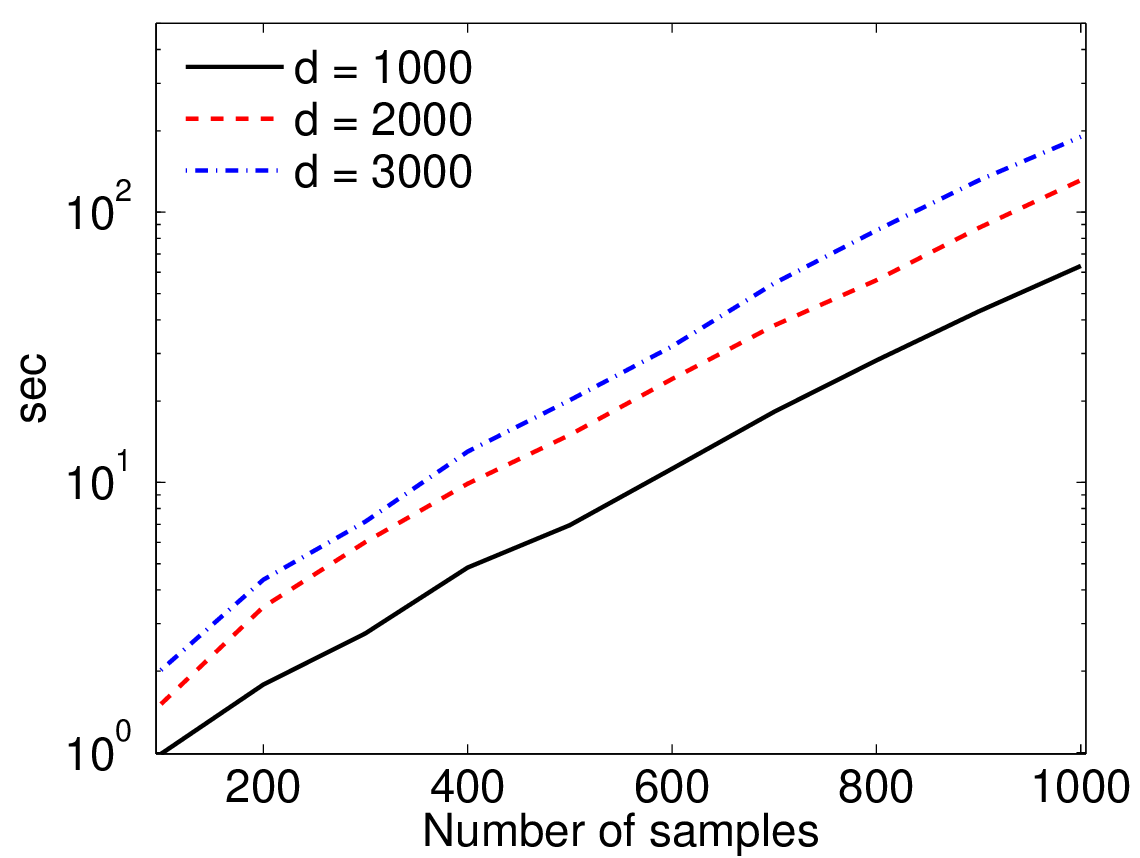}} \\ \vspace{-0.10cm}
(d) Computation time of HSIC Lasso for Data2
\end{minipage}
 \caption{(a),(b): Feature selection results for artificial datasets over 30 runs. The horizontal axis denotes the number of training samples, and the vertical axis denotes the fraction of correctly selected features. In HSIC Lasso and NOCCO Lasso, the regularization parameter $\lambda$ is set so that the number of non-zero coefficients is in $\{d^\ast, d^\ast+1, \ldots, d^\ast\ + 10\}$ where $d^\ast$ is the number of true features. Then, we use top $d^\ast$ features by ranking regression coefficients. In QPFS, FVM, and SpAM, we use top $d^\ast$ features by ranking coefficients. (c): Comparison of computation time for Data2. The horizontal axis denotes the number of entire features $d$, and the vertical axis denotes the computation time in log-scale. (d): Comparison of computation time for Data2 for HSIC Lasso. The horizontal axis denotes the number of training samples $n$, and the vertical axis denotes the computation time in log-scale. } 
    \label{fig:illustrative_example}
\end{center}
\vspace{-0.2in}
\end{figure*}

Figure~\ref{fig:illustrative_example} shows the feature selection accuracy of each method
over 30 runs as functions of the number of samples,
where the accuracy is measured by the fraction of correctly selected features
under the assumption that the number of true features is known.
As the figure clearly shows, the proposed HSIC Lasso and NOCCO
Lasso methods select good features in both additive and non-additive model cases.
SpAM also works very well for Data1, but it performs poorly for Data2
because the additivity assumption is violated in Data2.
QPFS and FVM behave similarly, but they tend to be outperformed by the proposed methods.

Next, we compare the computation time of each method. Here, we change the number of features in Data2 to $d=100, 200, \ldots, 1000$, while we fix the number of samples to $n = 100$.  Figure~\ref{fig:illustrative_example}-(c) shows the average computation time for each method over $30$ runs.  As can be observed, the computation time of HSIC Lasso and NOCCO Lasso increases mildly with respect to the number of features compared to that of SpAM, FVM, and QPFS. Figure~\ref{fig:illustrative_example}-(d) shows the average computation time of HSIC Lasso for $30$ runs. In this experiment, we use the table lookup trick for HSIC Lasso to deal with a relatively large number of samples. This shows that the lookup trick allows us to handle relatively large datasets.




\subsection{Real-World Datasets}
Next, we compare the performance of feature selection methods using real-world datasets.

\subsubsection{Multi-Class Classification}
We use four image datasets and two microarray datasets\footnote{\url{http://featureselection.asu.edu/datasets.php}}. Detailed information of the datasets is summarized in Table~\ref{tab:feat_data}. 

\begin{table*}[t]
  \centering
\caption{Summary of real-world datasets.}
\label{tab:feat_data}
\small
\begin{tabular}{c|l| c | c | c}
\hline
Type & Dataset & \# features ($d$) & \# samples ($n$) & \# Classes \\ \hline
Image & AR10P    & 2400    & 130  &  10 \\
& PIE10P   & 2400    & 210  &  10 \\
& PIX10P   & 10000   & 100  &  10 \\
& ORL10P   & 10000   & 100  &  10 \\ \hline
Microarray & TOX      & 5748    & 171  &   4 \\
& CLL-SUB  & 11340   & 111  &   3 \\
 \hline
\end{tabular}
\end{table*}

 In this experiment, we use 80$\%$ of samples for training and the rest for testing. We repeat the experiments 100 times by randomly shuffling training and test samples, and evaluate the performance of feature selection methods by the average classification accuracy. We use multi-class $\ell_2$-regularized kernel logistic regression (KLR) \citep{book:Hastie+Tibshirani+Friedman:2001,yamada2010semi}
with the Gaussian kernel for evaluating the classification accuracy when the top $m = 10, 20, \ldots, 50$ features selected by each method are used. In this paper, we first choose 50 features and then use top $m = 10,20,\ldots,50$ features having the largest absolute regression coefficients. In KLR, all tuning parameters 
such as the Gaussian width and the regularization parameter
are chosen based on 3-fold cross-validation. 

 We also investigate the  \emph{redundancy rate} (RED) \citep{AAAI:Zheng+etal:2010}\footnote{The original redundancy rate was defined with a plain correlation coefficient \citep{pearson1920notes}, not the absolute correlation coefficient \citep{AAAI:Zheng+etal:2010}. However, this is not appropriate as an error metric because negative correlation decreases RED. For this reason, we decided to use the absolute correlation coefficient.}, 
\begin{align*}
\textnormal{RED} = \frac{1}{m(m-1)} \sum_{\boldu_k, \boldu_j, k > l} |\rho_{k,l}|,
\end{align*}
 where $\rho_{k,l}$ is the correlation coefficient between the $k$-th and $l$-th features. A large RED score indicates that selected features are more strongly correlated to each other, that is, many redundant features are selected. 
Thus, as a feature selection method, a small redundancy rate is preferable.

\begin{figure*}[t!]
\begin{center}
\begin{minipage}[t]{0.325\linewidth}
\centering
  {\includegraphics[width=0.99\textwidth]{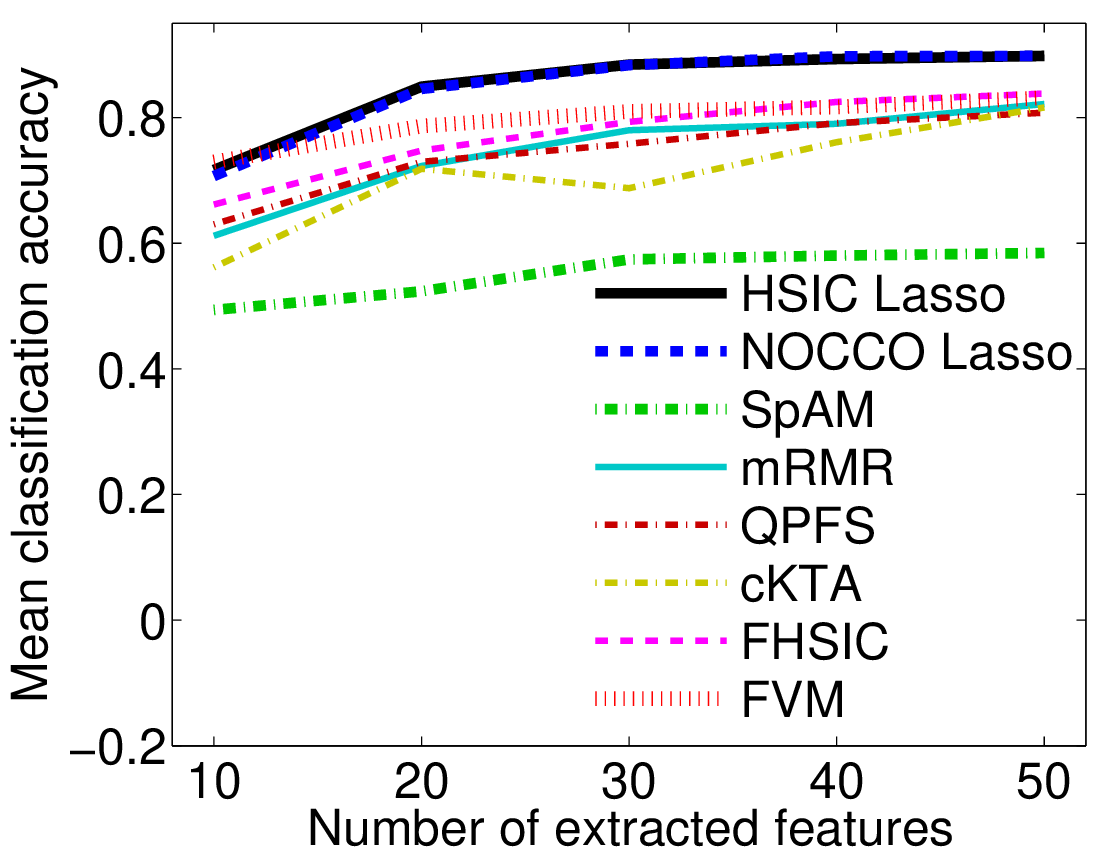}} \\ \vspace{-0.10cm}
(a) AR10P
\end{minipage}
\begin{minipage}[t]{0.325\linewidth}
\centering
  {\includegraphics[width=0.99\textwidth]{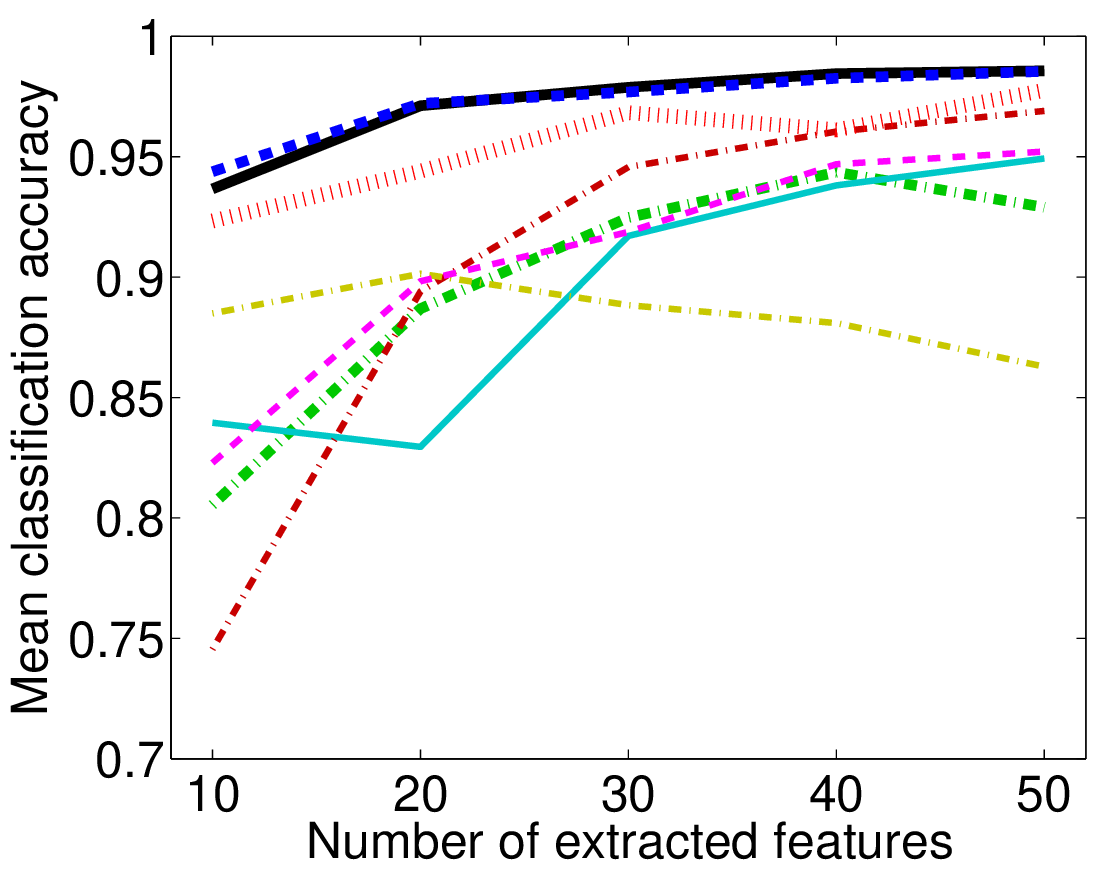}} \\ \vspace{-0.10cm}
(b) PIE10P
\end{minipage}
  \begin{minipage}[t]{0.325\linewidth}
\centering
{\includegraphics[width=0.99\textwidth]{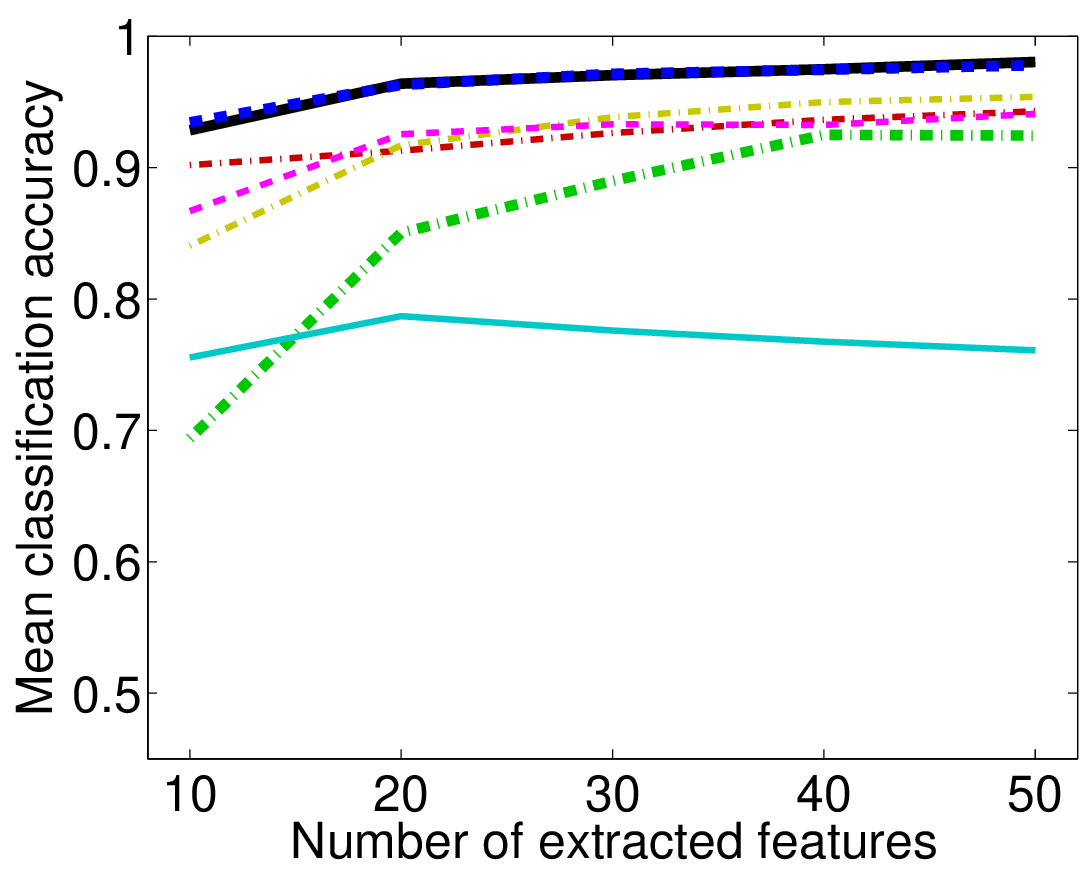}} \\  \vspace{-0.10cm}
(c) PIX10P
  \end{minipage} \\
\begin{minipage}[t]{0.325\linewidth}
\centering
  {\includegraphics[width=0.99\textwidth]{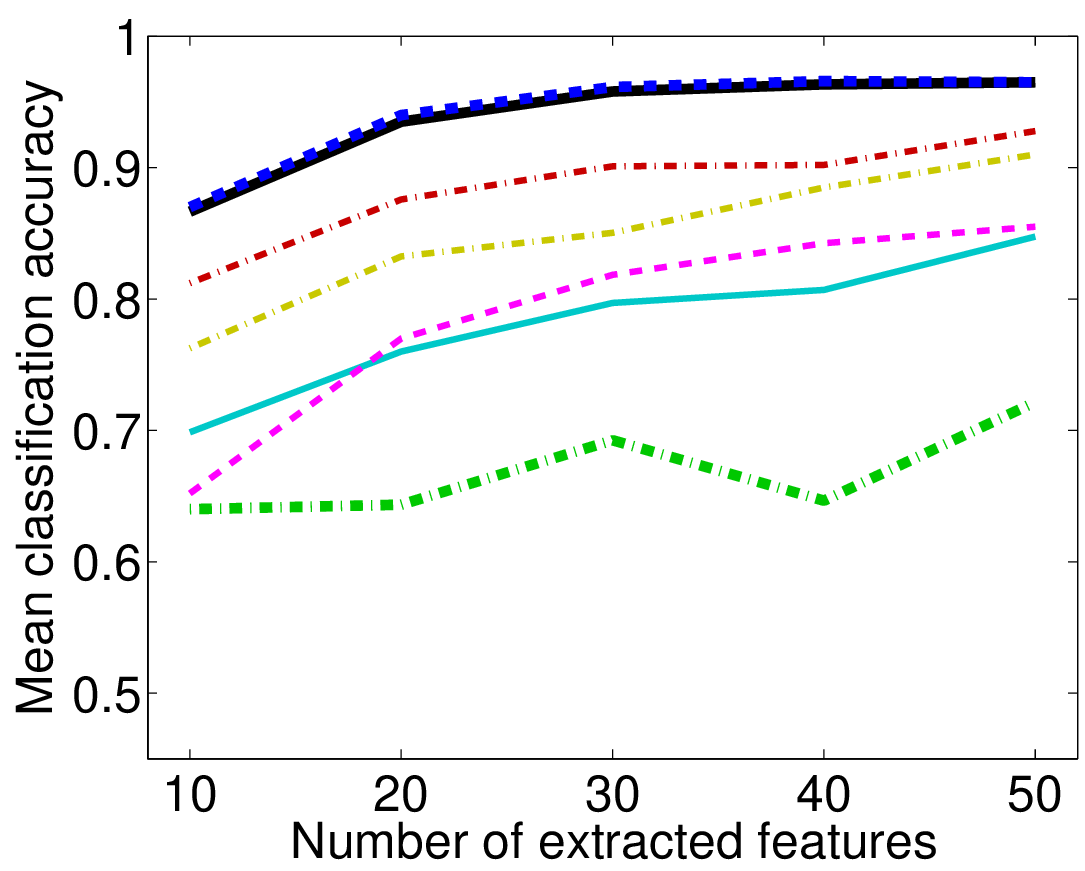}} \\ \vspace{-0.10cm}
(d) ORL10P
\end{minipage}
\begin{minipage}[t]{0.325\linewidth}
\centering
  {\includegraphics[width=0.99\textwidth]{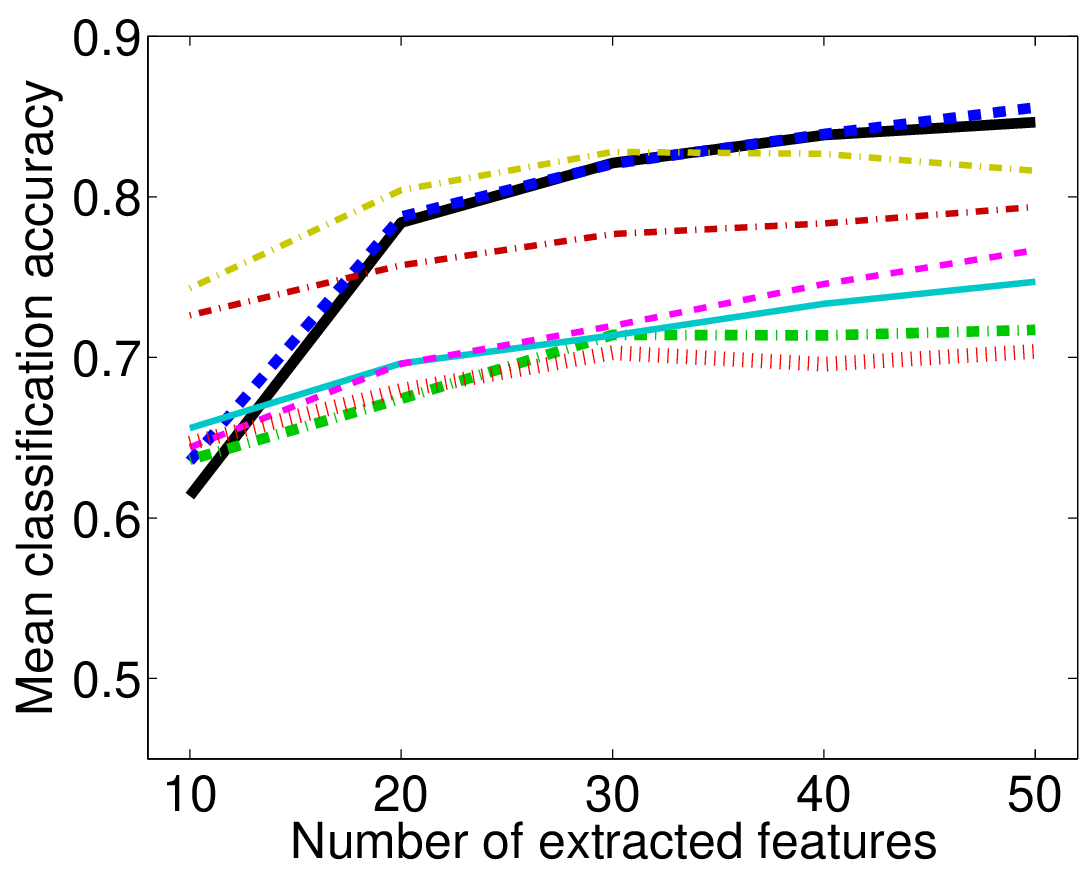}} \\ \vspace{-0.10cm}
(e) TOX
\end{minipage}
  \begin{minipage}[t]{0.325\linewidth}
\centering
{\includegraphics[width=0.99\textwidth]{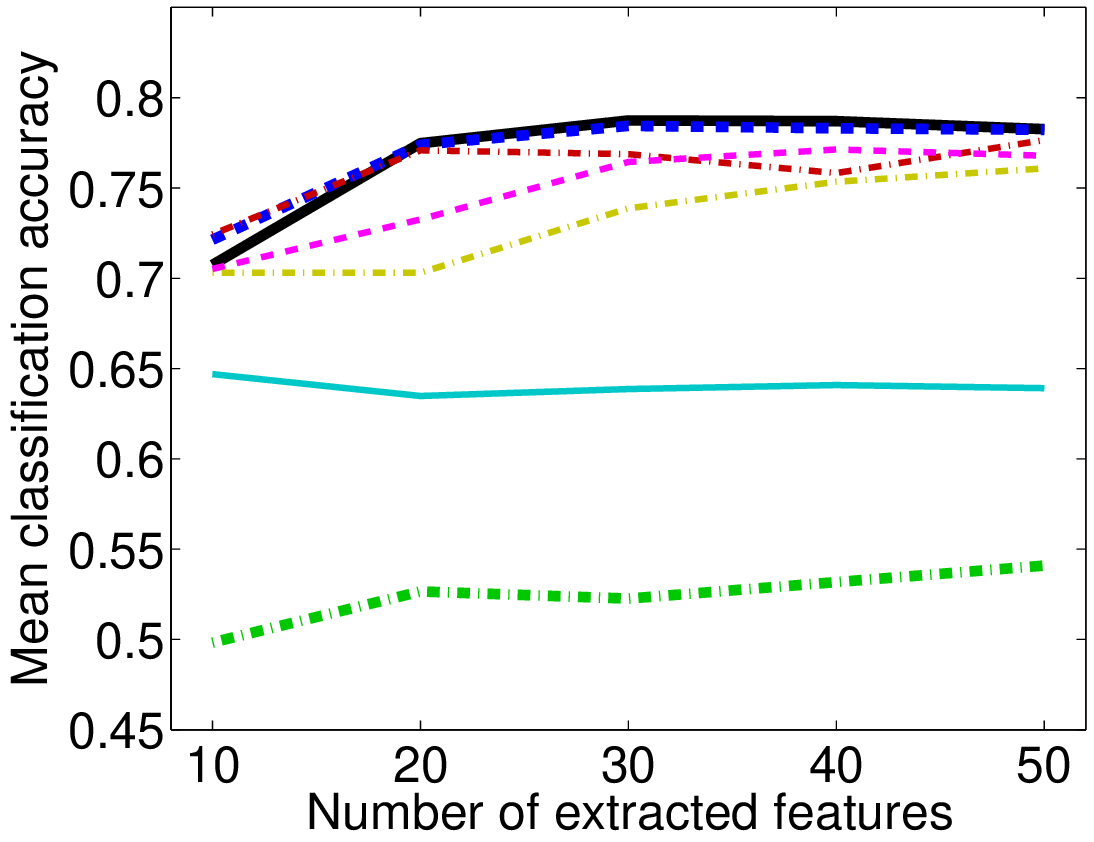}} \\  \vspace{-0.10cm}
(f) CLL-SUB
  \end{minipage}
 \caption{Mean classification accuracy for real-world data. The horizontal axis denotes the number of selected features, and the vertical axis denotes the mean classification accuracy.} 
    \label{fig:result_ASU}
\end{center}
\vspace{-0.2in}
\end{figure*}

\begin{table*}[t]
\centering
\caption{Mean classification accuracy (with standard deviations in
brackets) for real-world data. }
\label{table:classification_accuracy}
\begin{tabular}{c l@{\ }r@{\ }|l@{\ }r@{\ }|l@{\ }r@{\ }|l@{\ }r@{\ }|l@{\  }r@{\ }|l@{\ }r@{\ }|l@{\ }r@{\ }|l@{\ }r@{\ }}
\hline
\multicolumn{1}{c|}{\multirow{2}{*}{Dataset}}  & \multicolumn{2}{c|}{HSIC} & \multicolumn{2}{c|}{{NOCCO}} &  \multicolumn{2}{c|}{\multirow{2}{*}{SpAM}}  &  \multicolumn{2}{c|}{\multirow{2}{*}{FVM}}  & \multicolumn{2}{c|}{\multirow{2}{*}{mRMR}}  & \multicolumn{2}{c|}{\multirow{2}{*}{QPFS}} & \multicolumn{2}{c|}{\multirow{2}{*}{cKTA}} & \multicolumn{2}{c}{\multirow{2}{*}{FHSIC}}\\
\multicolumn{1}{c|}{}  & \multicolumn{2}{c|}{Lasso} & \multicolumn{2}{c|}{{Lasso}}  &  \multicolumn{2}{c|}{} & \multicolumn{2}{c|}{} & \multicolumn{2}{c|}{} & \multicolumn{2}{c|}{}  & \multicolumn{2}{c|}{} & \multicolumn{2}{c}{}\\
\hline  
\multicolumn{1}{l|}{AR10P}    & {\bf .848}         & (.111)   & .846    & (.111) & .551 & (.109) & .795 & (.121) & .745 & (.136) & .743 & (.137) & .709 & (.207) & .773 & (.122) \\
\multicolumn{1}{l|}{PIE10P}   & .971  & (.032)   & {\bf .972}          & (.031) & .898 & (.109) & .955 & (.062) & .895 & (.118) & .952 & (.067) & .884 & (.218) & .908 & (.091)  \\
\multicolumn{1}{l|}{PIX10P}   & {\bf .964}   & (.043)   & {\bf .964} & (.042) & .857 & (.146) & ---  & (---)  & .769 & (.124) & .924 & (.067) & .920 & (.090) & .920 & (.100) \\
\multicolumn{1}{l|}{ORL10P}   & .938   & (.068)   & {\bf .941} & (.066) & .669 & (.120) & ---  & (---)  & .782 & (.138) & .884 & (.096) & .848 & (.140) & .788 & (.132)  \\
\multicolumn{1}{l|}{TOX}      & .781         & (.119)   & .788    & (.113) & .691 & (.087) & .686 & (.085) & .709 & (.084) & .769 & (.077) & {\bf .804} & (.110) & .715 & (.087) \\
\multicolumn{1}{l|}{CLL-SUB}  & .768         & (.087)   & {\bf .769}          & (.084) & .524 & (.112) & ---  & (---)  & .640 & (.098) & .760 & (.116) & .709 & (.104) & .732 & (.141) \\
 \hline
\end{tabular}
\end{table*}

\begin{table*}[t]
\centering
\caption{Mean redundancy rate (with standard deviations in
brackets) for real-world data. }
\label{table:redundancy_rate}
\begin{tabular}{c l@{\ }r@{\ }|l@{\ }r@{\ }|l@{\ }r@{\ }|l@{\ }r@{\ }|l@{\  }r@{\ }|l@{\ }r@{\ }|l@{\ }r@{\ }|l@{\ }r@{\ }}
\hline
\multicolumn{1}{c|}{\multirow{2}{*}{Dataset}}  & \multicolumn{2}{c|}{HSIC} & \multicolumn{2}{c|}{{NOCCO}} &  \multicolumn{2}{c|}{\multirow{2}{*}{SpAM}}  &  \multicolumn{2}{c|}{\multirow{2}{*}{FVM}}  & \multicolumn{2}{c|}{\multirow{2}{*}{mRMR}}  & \multicolumn{2}{c|}{\multirow{2}{*}{QPFS}} & \multicolumn{2}{c|}{\multirow{2}{*}{cKTA}} & \multicolumn{2}{c}{\multirow{2}{*}{FHSIC}}\\
\multicolumn{1}{c|}{}  & \multicolumn{2}{c|}{Lasso} & \multicolumn{2}{c|}{{Lasso}}  &  \multicolumn{2}{c|}{} & \multicolumn{2}{c|}{} & \multicolumn{2}{c|}{} & \multicolumn{2}{c|}{}  & \multicolumn{2}{c|}{} & \multicolumn{2}{c}{}\\
\hline  
\multicolumn{1}{l|}{AR10P}    & .196  & (.028)   & {\bf .195}       & (.028) & .255 & (.036) & .260 & (.039) & .268 & (.038) & .217 & (.050) & .235 & (.034) & .350 & (.091) \\
\multicolumn{1}{l|}{PIE10P}   &  {\bf .135}  & (.014)  & .139       & (.017) & .250 & (.042) & .193 & (.029) & .225 & (.036) & .183 & (.026) & .155 & (.021) & .285 & (.059)  \\
\multicolumn{1}{l|}{PIX10P}   & {\bf .177}  & (.023)   & {\bf .174} & (.023) & .388 & (.105) & ---  & (---)  & .200 & (.066) & .286 & (.057) & .198 & (.036) & .348 & (.064) \\
\multicolumn{1}{l|}{ORL10P}   & {\bf .192}  & (.026)   & {\bf .191} & (.025) & .300 & (.047) & ---  & (---)  & .294 & (.095) & .204 & (.032) & .191 & (.034) & .225 & (.045)  \\
\multicolumn{1}{l|}{TOX}      & .382        & (.027)   & .381 & (.027) & .391 & (.028) & .422 & (.031) & .386 & (.032) & .384 & (.028) & {\bf .371} & (.040) & .396 & (.036) \\
\multicolumn{1}{l|}{CLL-SUB}  & .344        & (.034)   & .345 & (.034) & .403 & (.058) & ---  & (---)  & .328 & (.039) & .322 & (.033) & {\bf .281} & (.050) & .352 & (.061) \\
 \hline
\end{tabular}
\end{table*}

\begin{figure*}[t!]
\begin{center}
\begin{minipage}[t]{0.325\linewidth}
\centering
  {\includegraphics[width=0.99\textwidth]{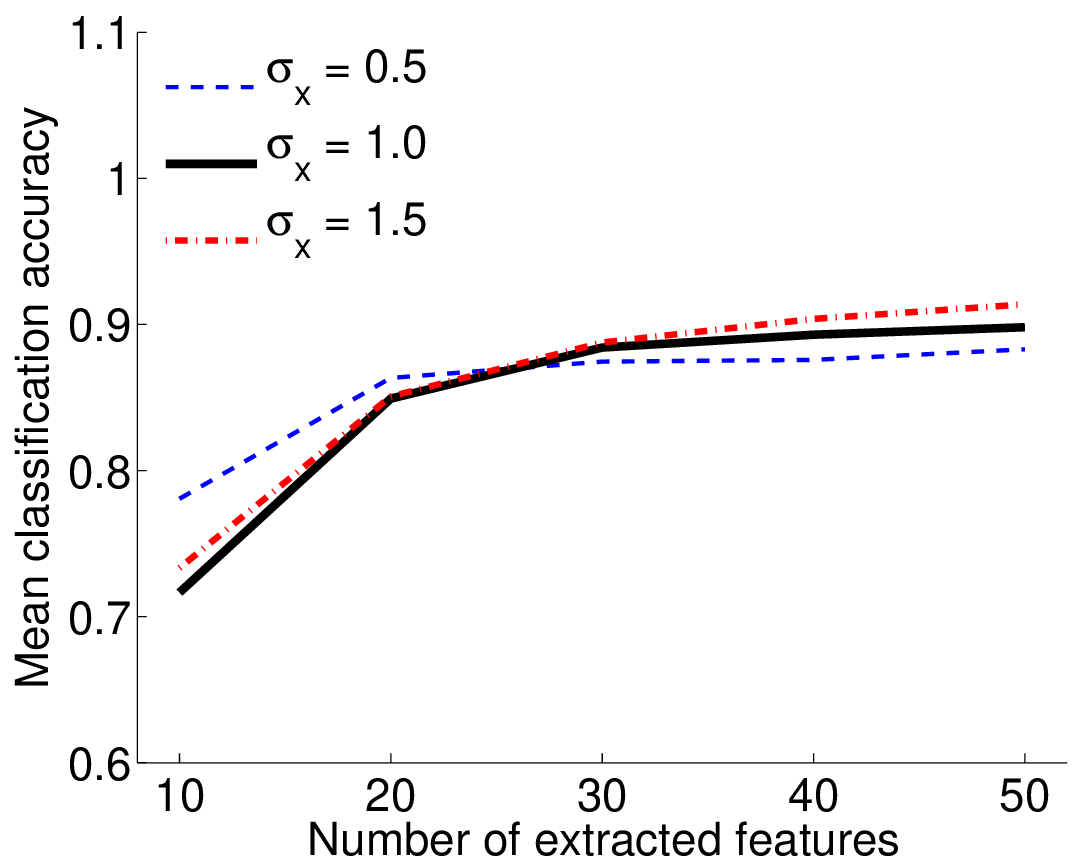}} \\ \vspace{-0.10cm}
(a) AR10P
\end{minipage}
\begin{minipage}[t]{0.325\linewidth}
\centering
  {\includegraphics[width=0.99\textwidth]{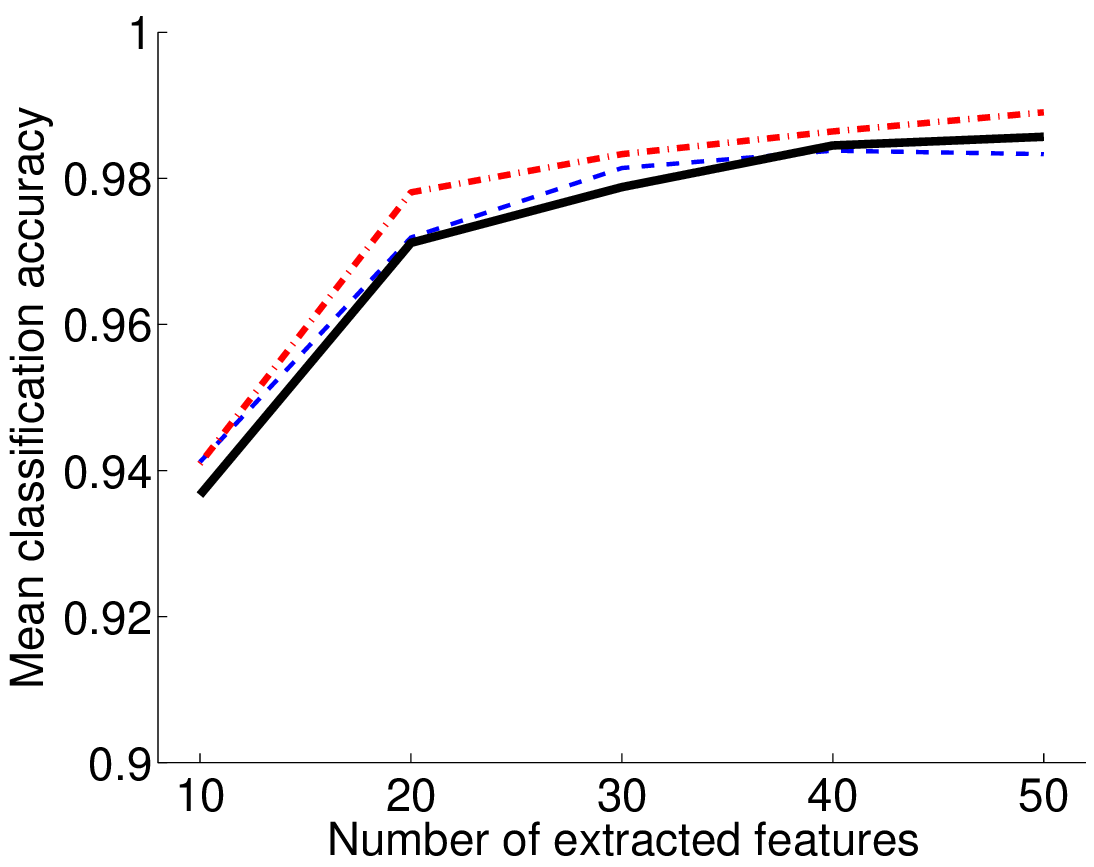}} \\ \vspace{-0.10cm}
(b) PIE10P
\end{minipage}
  \begin{minipage}[t]{0.325\linewidth}
\centering
{\includegraphics[width=0.99\textwidth]{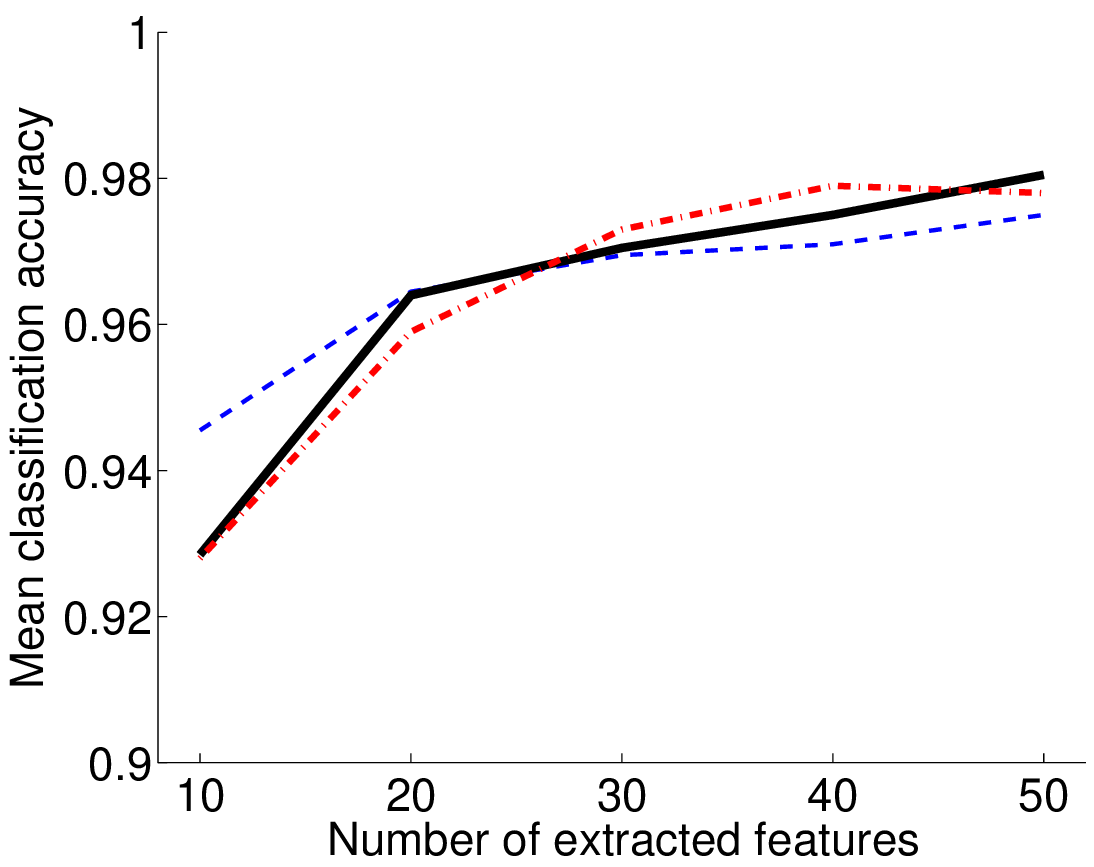}} \\  \vspace{-0.10cm}
(c) PIX10P
  \end{minipage} \\
\begin{minipage}[t]{0.325\linewidth}
\centering
  {\includegraphics[width=0.99\textwidth]{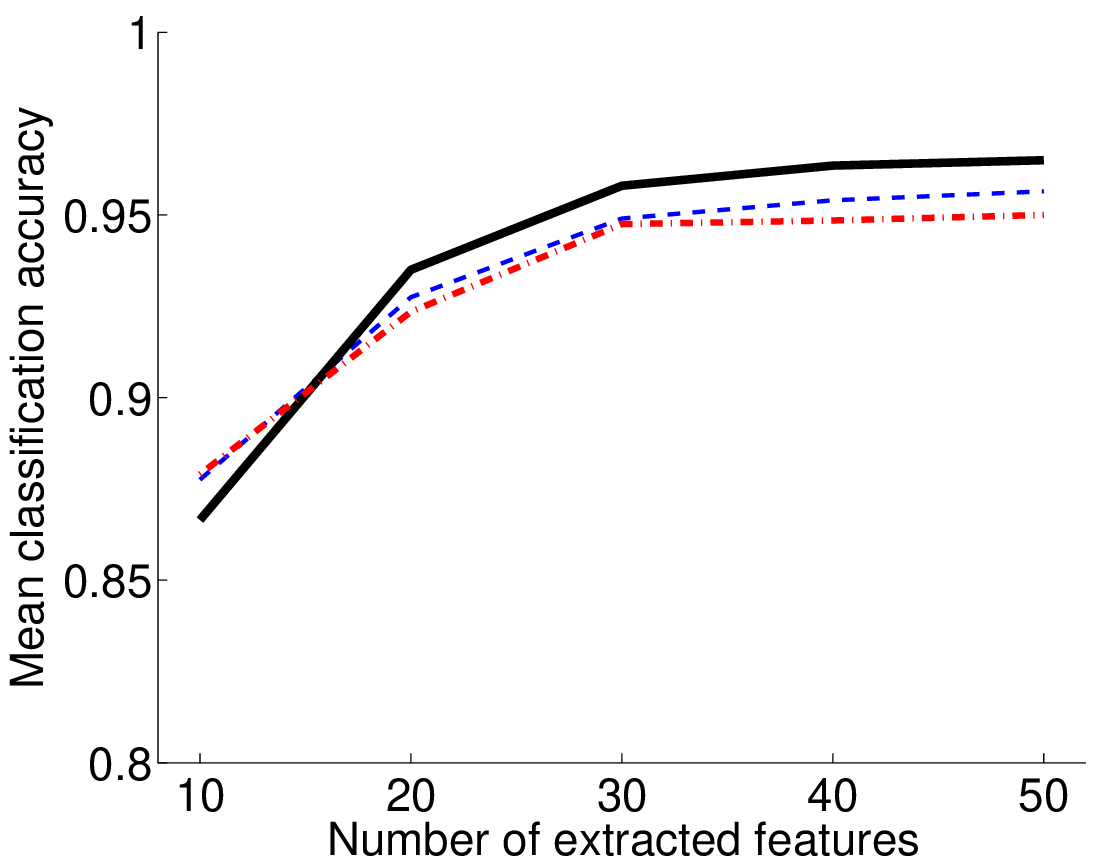}} \\ \vspace{-0.10cm}
(d) ORL10P
\end{minipage}
\begin{minipage}[t]{0.325\linewidth}
\centering
  {\includegraphics[width=0.99\textwidth]{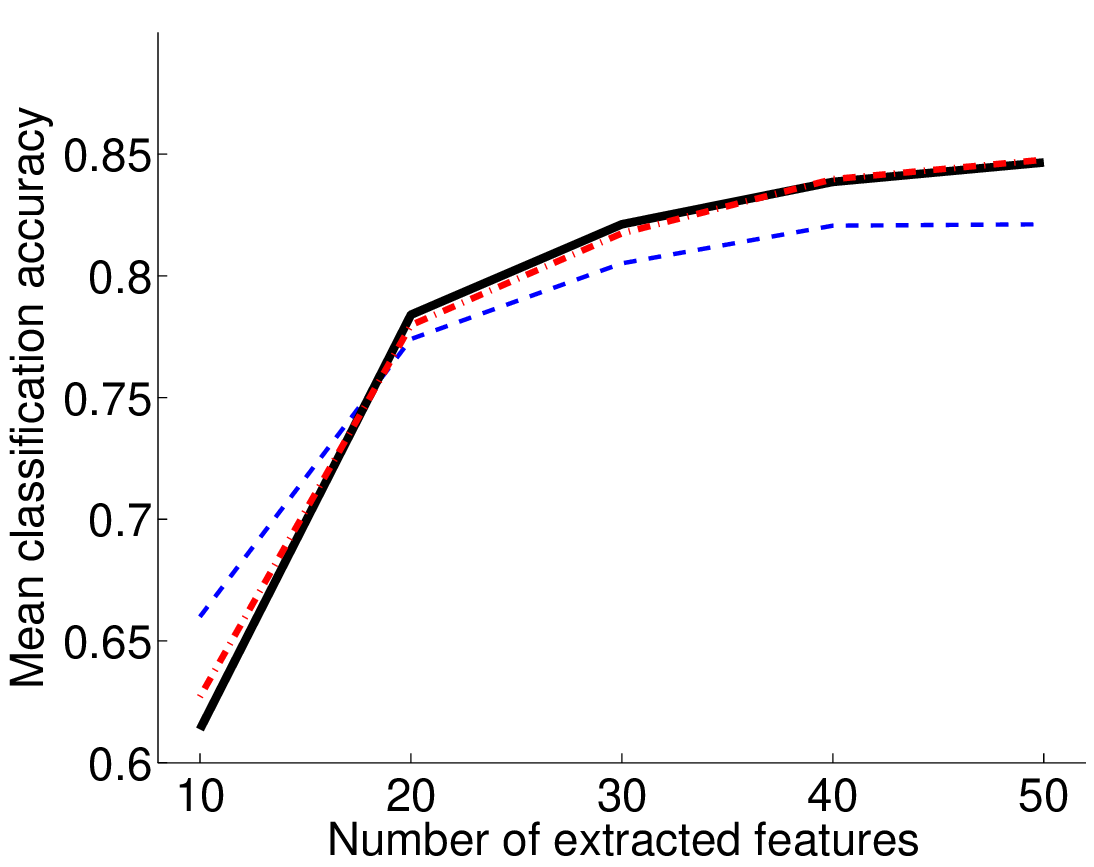}} \\ \vspace{-0.10cm}
(e) TOX
\end{minipage}
  \begin{minipage}[t]{0.325\linewidth}
\centering
{\includegraphics[width=0.99\textwidth]{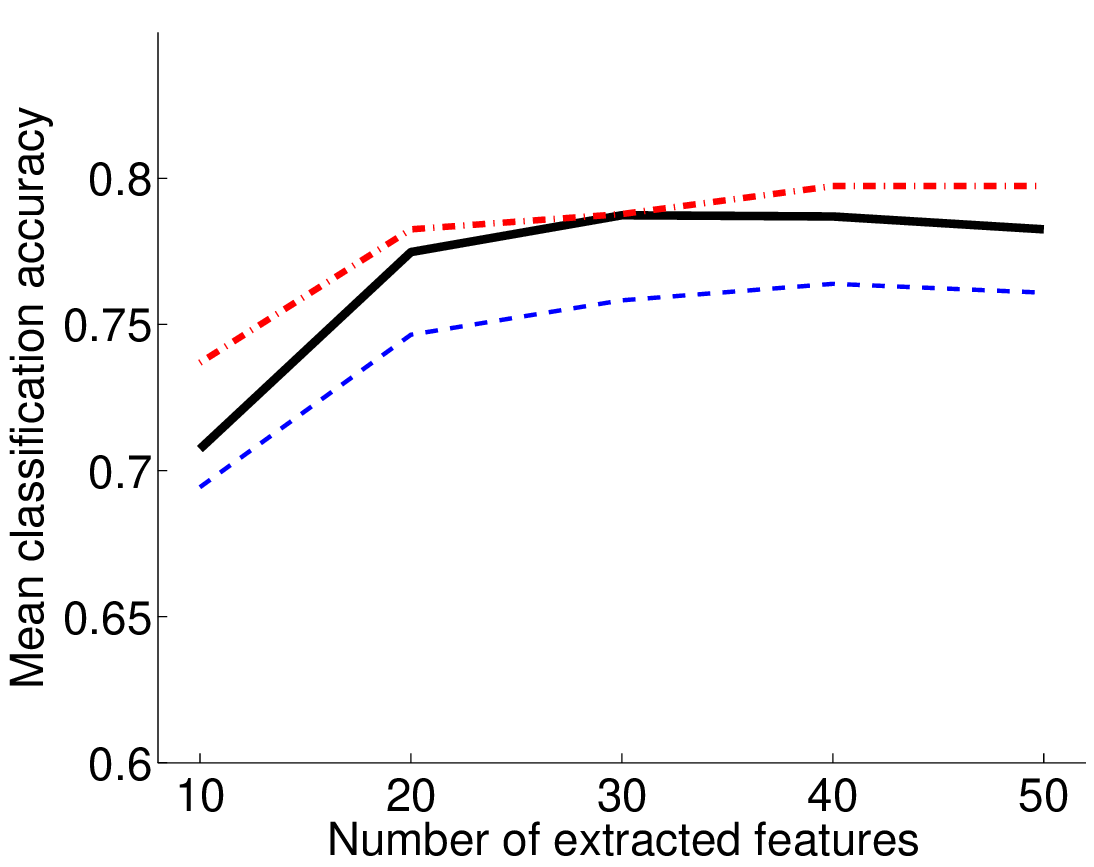}} \\  \vspace{-0.10cm}
(f) CLL-SUB
  \end{minipage}
 \caption{Mean classification accuracy of HSIC Lasso with different Gaussian widths. The horizontal axis denotes the number of selected features, and the vertical axis denotes the mean classification accuracy.} 
    \label{fig:result_ASU_HSICLasso_width}
\end{center}
\vspace{-0.2in}
\end{figure*}

\begin{figure*}[t!]
\begin{center}
\begin{minipage}[t]{0.325\linewidth}
\centering
  {\includegraphics[width=0.99\textwidth]{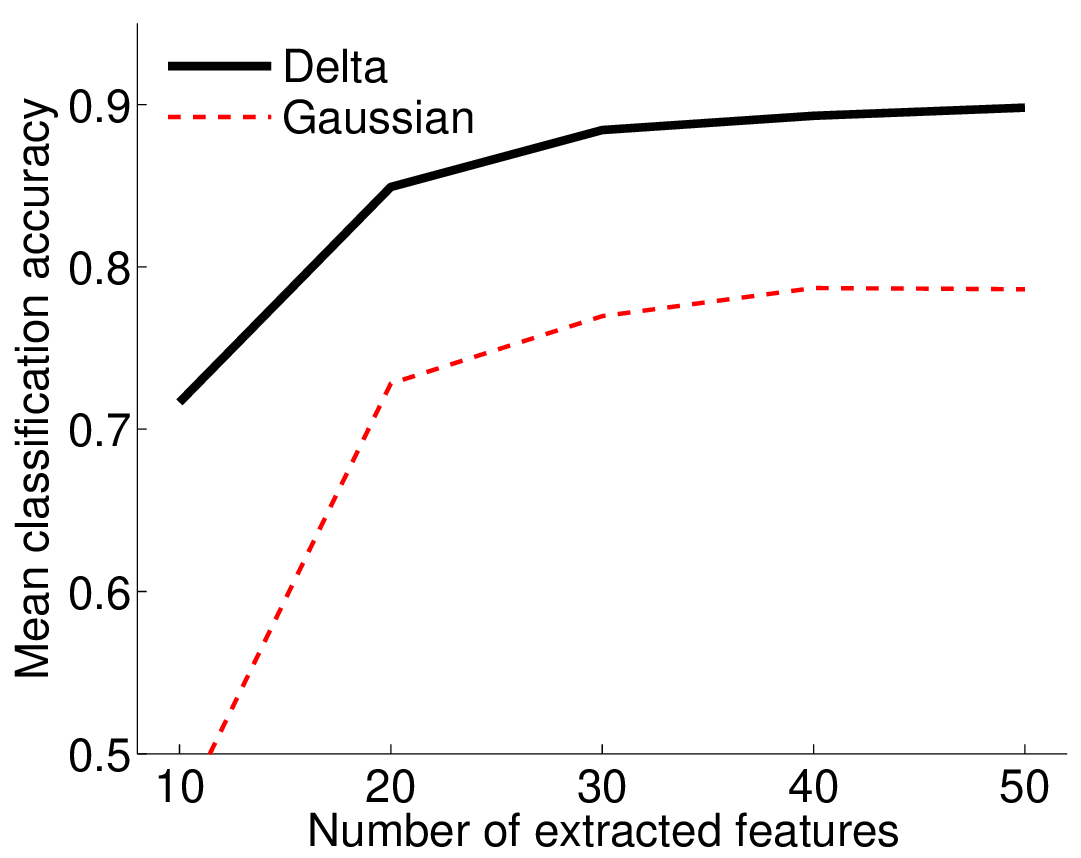}} \\ \vspace{-0.10cm}
(a) AR10P
\end{minipage}
\begin{minipage}[t]{0.325\linewidth}
\centering
  {\includegraphics[width=0.99\textwidth]{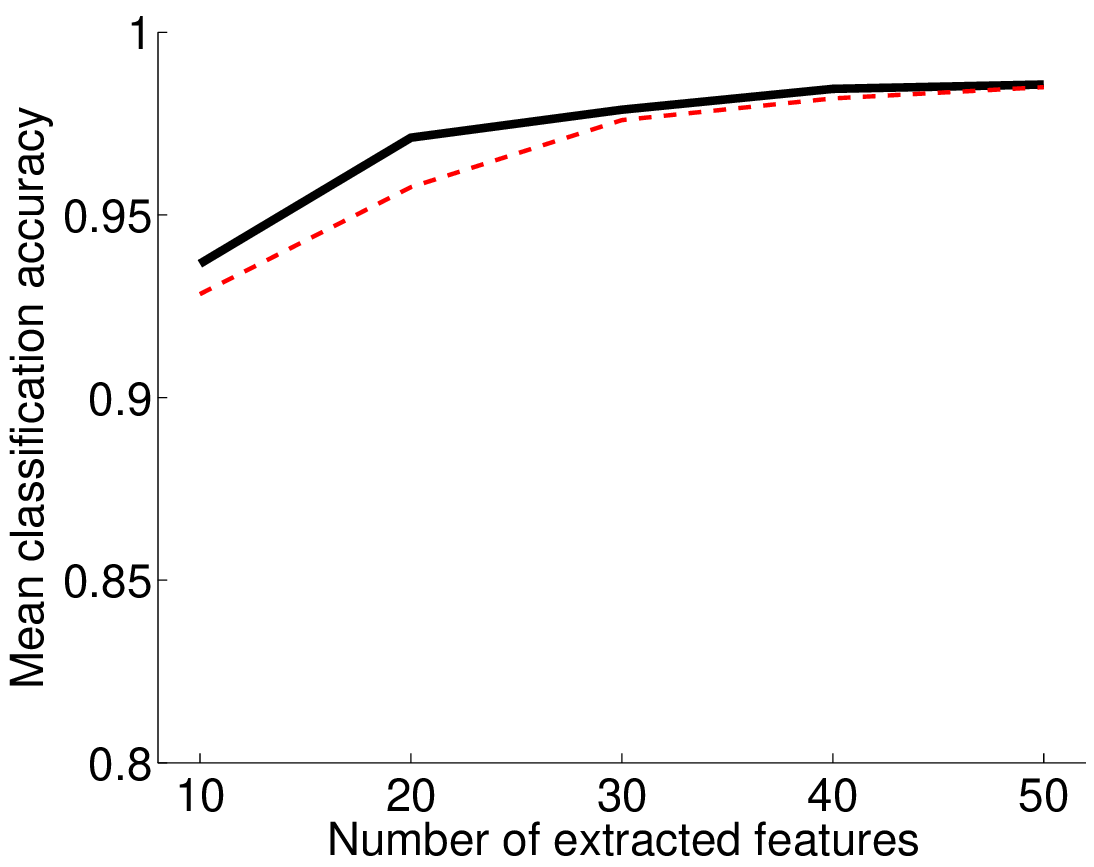}} \\ \vspace{-0.10cm}
(b) PIE10P
\end{minipage}
  \begin{minipage}[t]{0.325\linewidth}
\centering
{\includegraphics[width=0.99\textwidth]{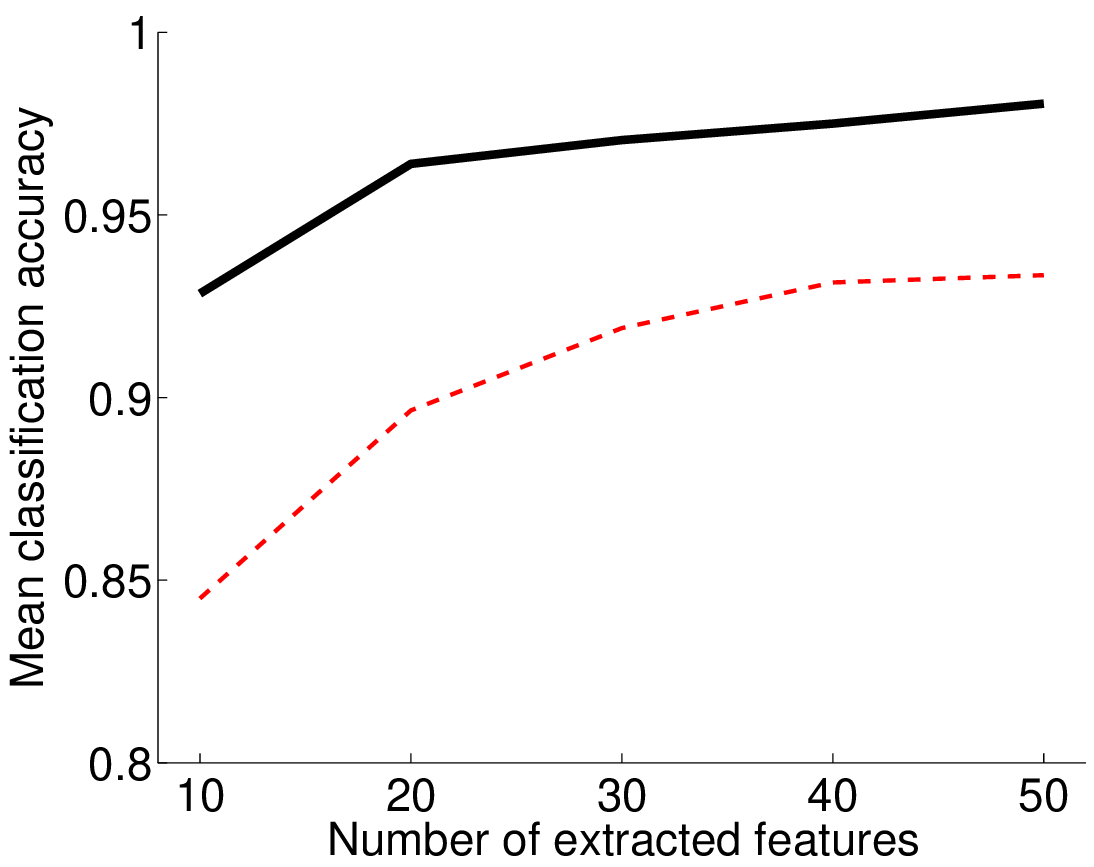}} \\  \vspace{-0.10cm}
(c) PIX10P
  \end{minipage} \\
\begin{minipage}[t]{0.325\linewidth}
\centering
  {\includegraphics[width=0.99\textwidth]{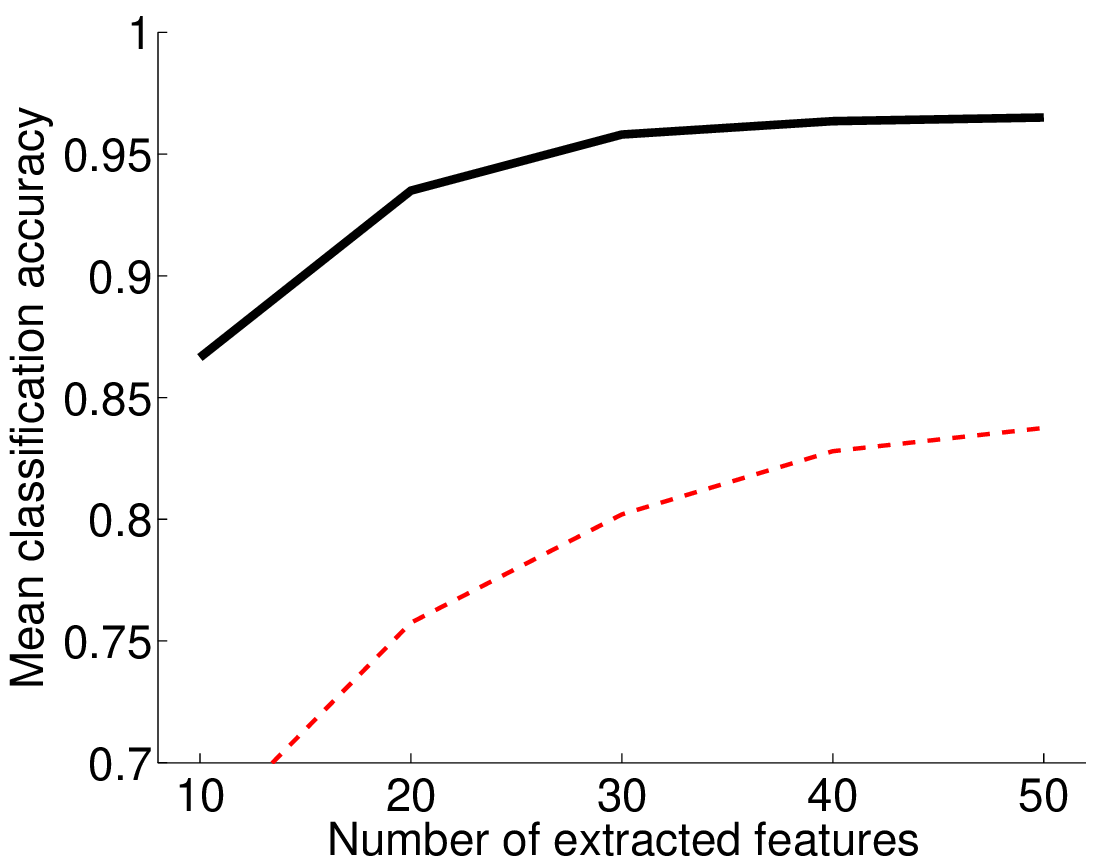}} \\ \vspace{-0.10cm}
(d) ORL10P
\end{minipage}
\begin{minipage}[t]{0.325\linewidth}
\centering
  {\includegraphics[width=0.99\textwidth]{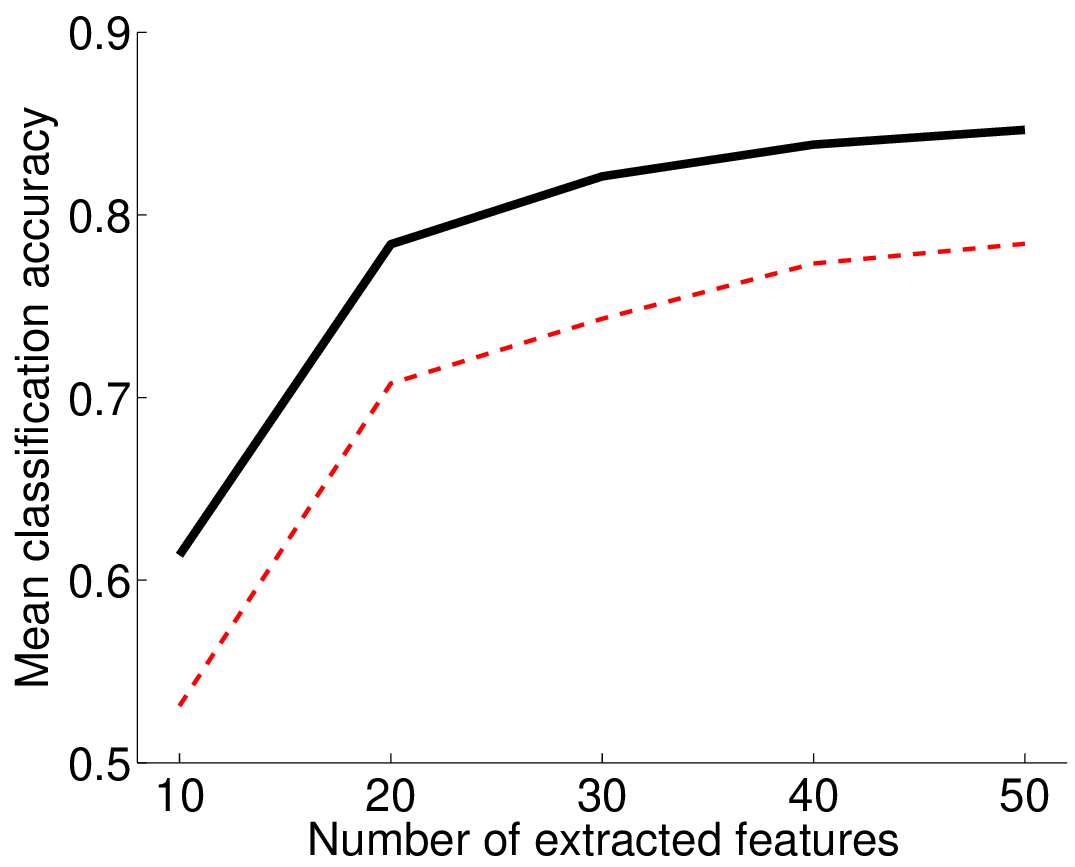}} \\ \vspace{-0.10cm}
(e) TOX
\end{minipage}
  \begin{minipage}[t]{0.325\linewidth}
\centering
{\includegraphics[width=0.99\textwidth]{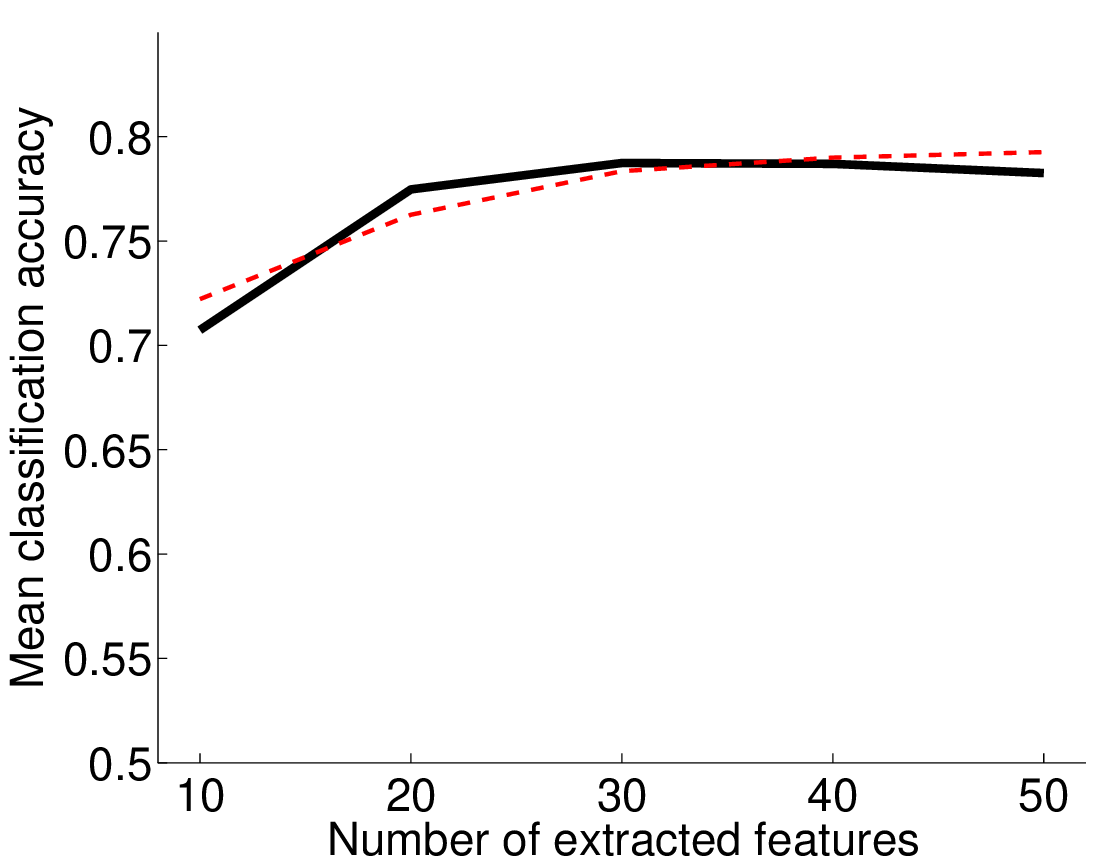}} \\  \vspace{-0.10cm}
(f) CLL-SUB
  \end{minipage}
 \caption{Mean classification accuracy of HSIC Lasso with different output kernels. Here, we use the delta kernel and the Gaussian kernel. The horizontal axis denotes the number of selected features, and the vertical axis denotes the mean classification accuracy.} 
    \label{fig:result_ASU_HSICLasso_outkernel}
\end{center}
\vspace{-0.2in}
\end{figure*}

\vspace{0.1in}
\noindent{\bf Results:}
Figure~\ref{fig:result_ASU} shows the mean classification accuracy over 100 runs
as functions of the number of selected features. Table~\ref{table:classification_accuracy} shows the average classification accuracy rates for the top $m=50$ features selected by each method. In this experiment, since the computation cost of FVM was too high for datasets with a large number of features, we only included the FVM results for the datasets with a small number of features (i.e., AR10P, PIE10P, and TOX).
The graphs in Figure~\ref{fig:result_ASU} clearly show that HSIC Lasso and NOCCO Lasso compare favorably with existing methods for the image datasets (i.e., AR10P, PIE10P, PIX10P, and ORL10P) in terms of the classification accuracy, and they are comparable to existing methods for the microarray datasets (i.e., TOX and CLL-SUB). 

Table~\ref{table:redundancy_rate} shows the RED values for the top $m=50$ features selected by each method. 
As can be observed, HSIC Lasso and NOCCO Lasso tend to have smaller RED values, and thus they select less redundant features.

\vspace{0.1in}
\noindent{\bf Role of the Gaussian Width and the Output Kernel:} In the proposed methods, the Gaussian width and the output kernel $L(y,y')$ must be chosen manually. We carried out a set of experiments to show the sensitivity of choosing the Gaussian width and the output kernel in Figures~\ref{fig:result_ASU_HSICLasso_width} and \ref{fig:result_ASU_HSICLasso_outkernel}. Note that, since the performance of HSIC Lasso and NOCCO Lasso are comparable, we here only evaluate HSIC Lasso. As can be seen in Figure \ref{fig:result_ASU_HSICLasso_width}, the proposed method is not so sensitive to the Gaussian width. From the output kernel comparison in Figure \ref{fig:result_ASU_HSICLasso_outkernel}, we found that HSIC Lasso with delta kernel clearly outperforms that with Gaussian kernel. Thus using different input and output kernels is important for feature selection in classification scenarios.


\subsubsection{High-Dimensional Regression}
We also evaluate our proposed method with the Affymetric GeneChip Rat Genome 230 2.0 Array data set \citep{scheetz2006regulation}. The data set consists of 120 rat subjects with 31098 genes which were measured from eye tissue. Similar to \citet{huang2010variable}, we focus on finding genes that are related to the TRIM32 gene, which was recently found to cause the Bardet-Biedl syndrome. Note that, since TRIM32 takes real values, this is a regression problem.

In this experiment, we use 80$\%$ of samples for training and the rest for testing. We repeat the experiments 100 times by randomly shuffling training and test samples, and evaluate the performance of feature selection methods by the mean squared error. In addition, we use the correlation coefficient between the predicted and the true TRIM32 values, which is a popular performance metric in biology community. We use  kernel regression (KR) \citep{book:Schoelkopf+Smola:2002} with the Gaussian kernel for evaluating the mean squared error and the mean correlation when the top $m = 10, 20, \ldots, 50$ features selected by each method are used. We first choose 50 features and then use top $m = 10,20,\ldots,50$ features having the largest absolute regression coefficients. In KR, all tuning parameters  such as the Gaussian width and the regularization parameter are chosen based on 3-fold cross-validation. 

\begin{figure*}[t!]
\begin{center}
\begin{minipage}[t]{0.32\linewidth}
\centering
  {\includegraphics[width=0.99\textwidth]{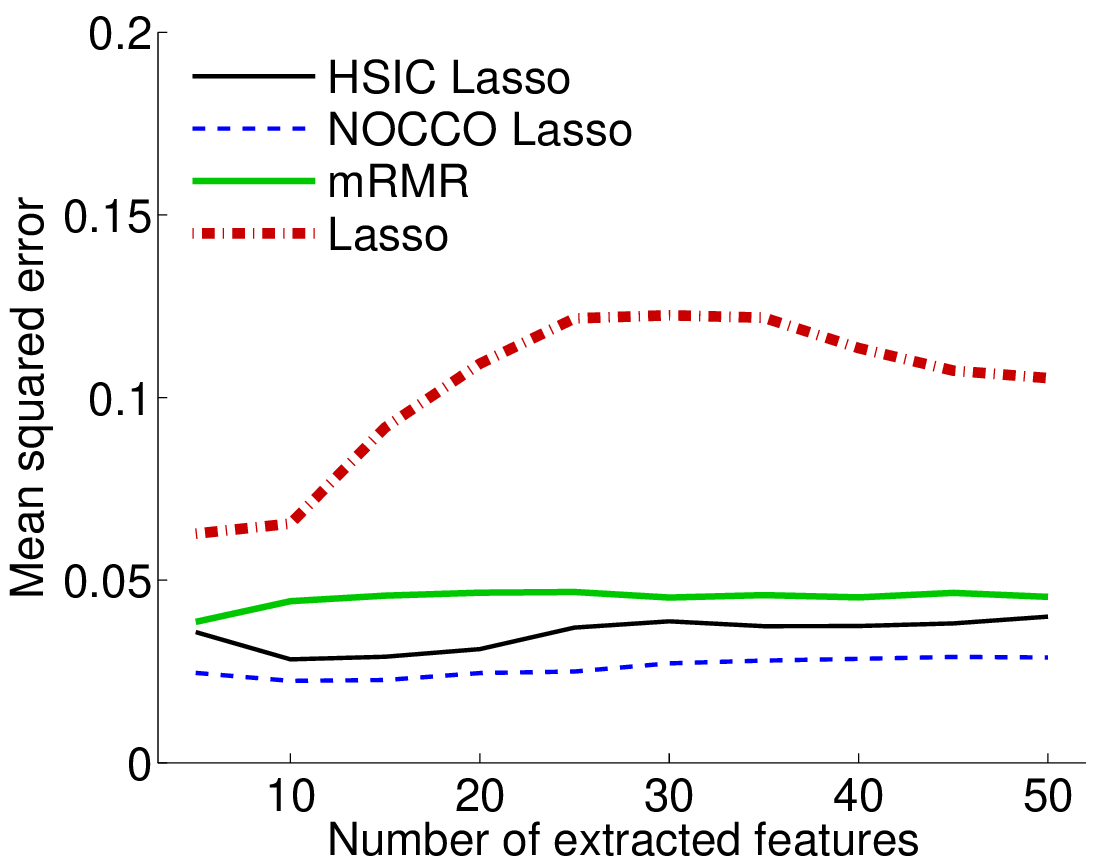}} \\ \vspace{-0.10cm}
(a) MSE
\end{minipage}
\begin{minipage}[t]{0.32\linewidth}
\centering
  {\includegraphics[width=0.99\textwidth]{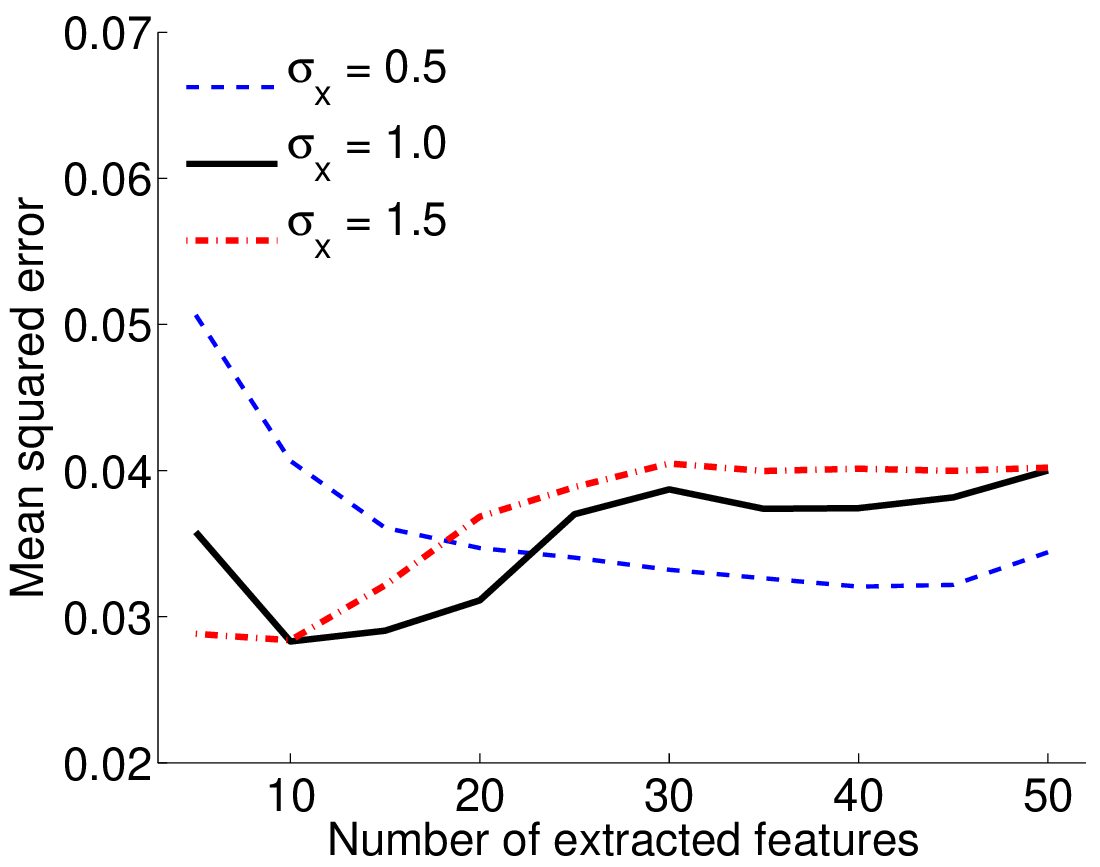}} \\ \vspace{-0.10cm}
(b) MSE (HSIC Lasso)
\end{minipage}
\begin{minipage}[t]{0.32\linewidth}
\centering
  {\includegraphics[width=0.99\textwidth]{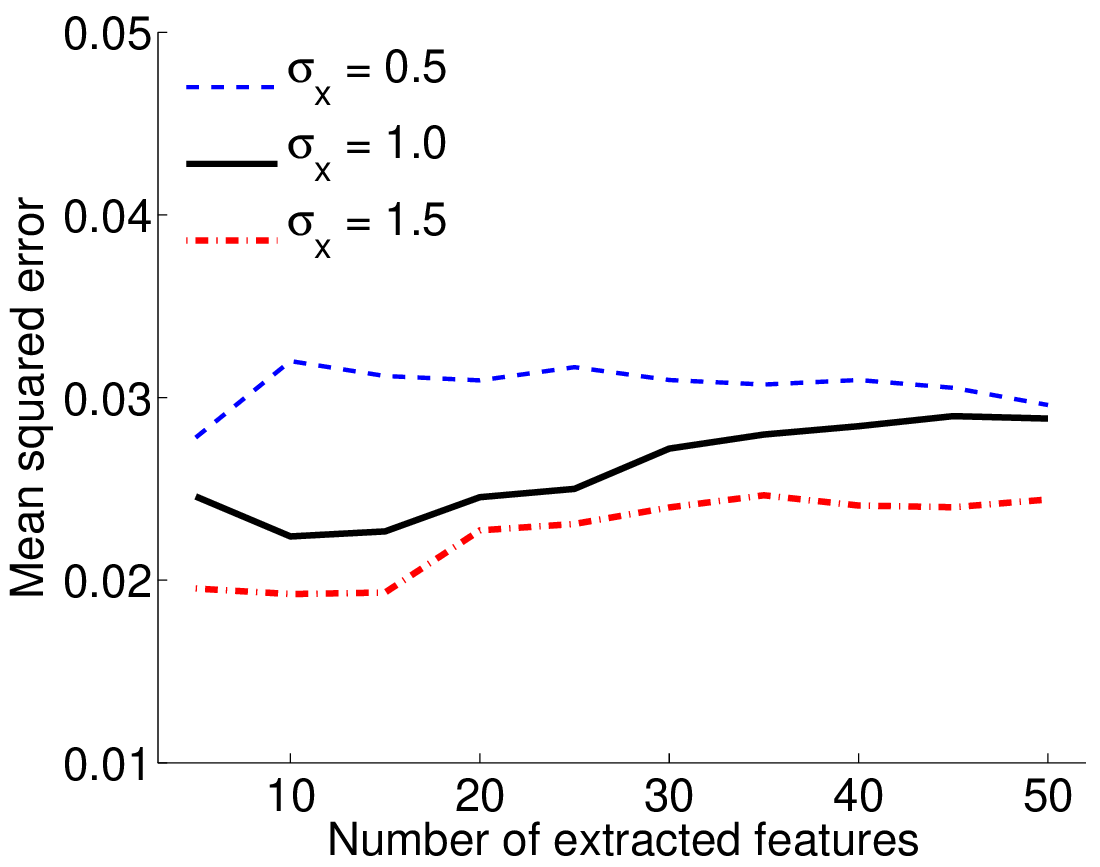}} \\ \vspace{-0.10cm}
(c) MSE (NOCCO Lasso)
\end{minipage}
  \begin{minipage}[t]{0.32\linewidth}
\centering
{\includegraphics[width=0.99\textwidth]{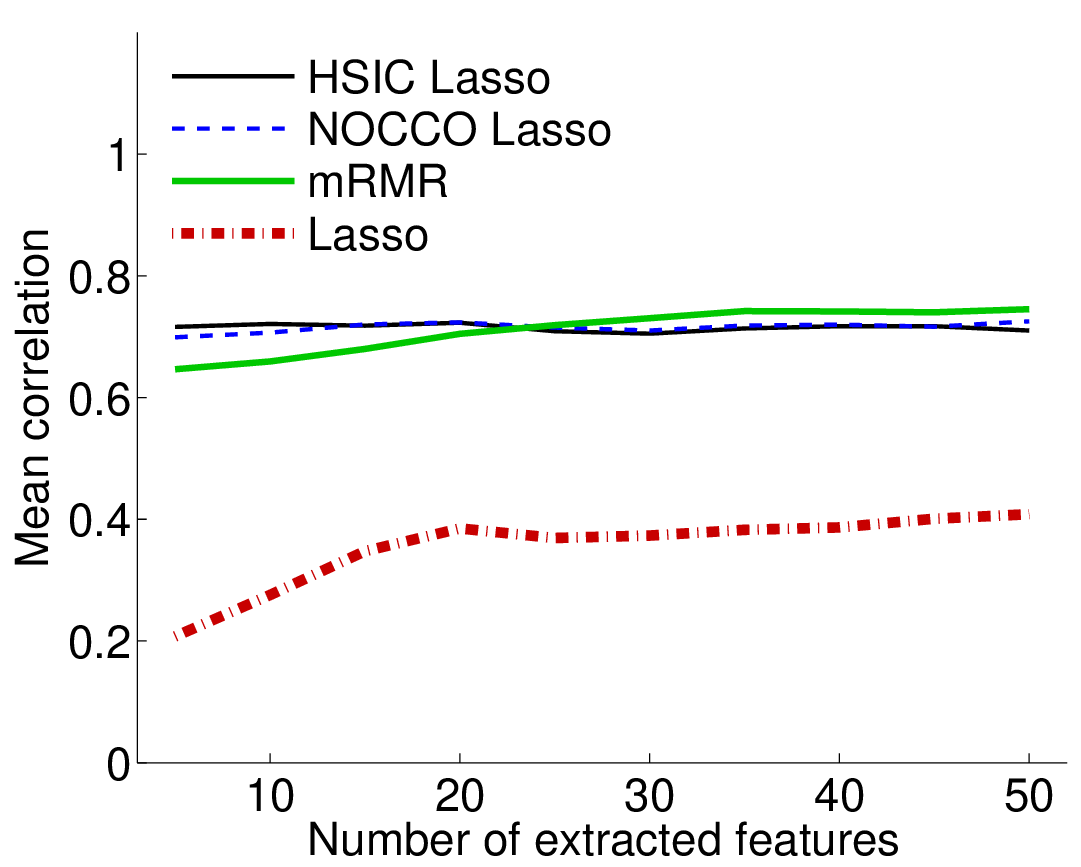}} \\  \vspace{-0.10cm}
(d) Mean correlation coefficient
  \end{minipage} 
\begin{minipage}[t]{0.32\linewidth}
\centering
  {\includegraphics[width=0.99\textwidth]{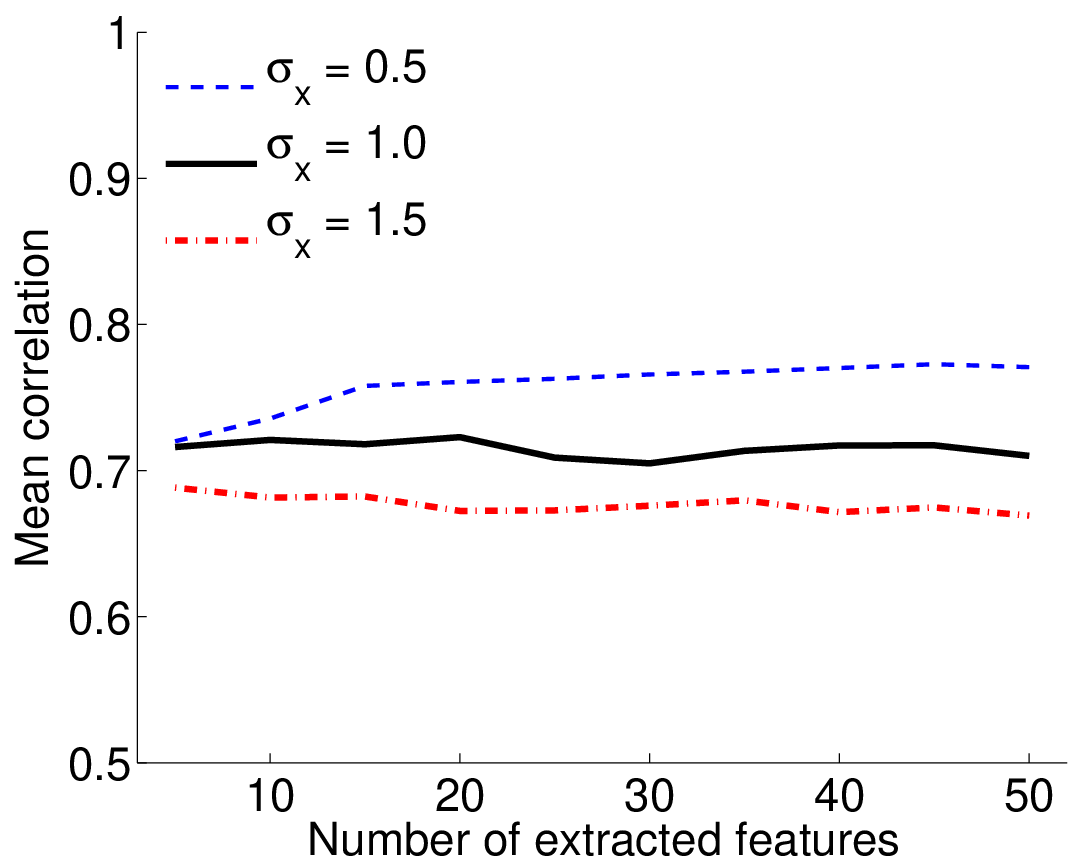}} \\ \vspace{-0.10cm}
(e) Mean correlation coefficient (HSIC Lasso)
\end{minipage}
\begin{minipage}[t]{0.32\linewidth}
\centering
  {\includegraphics[width=0.99\textwidth]{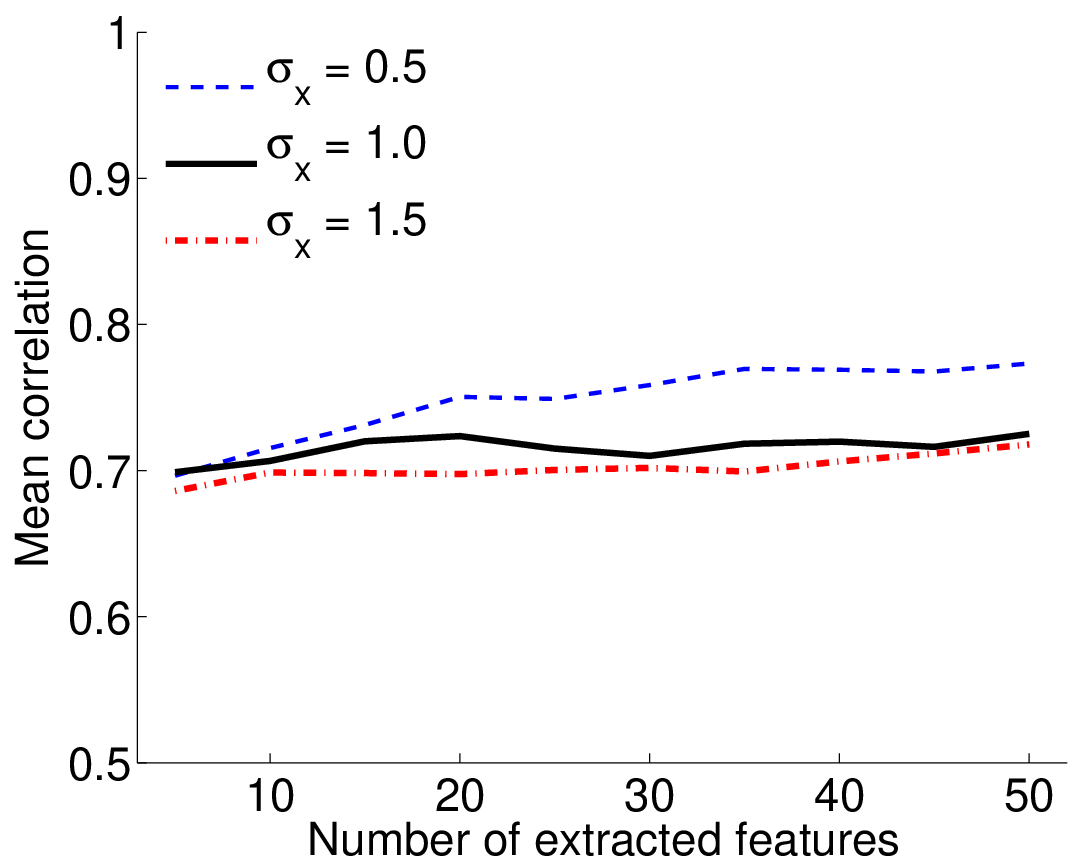}} \\ \vspace{-0.10cm}
(f) Mean correlation coefficient (NOCCO Lasso)
\end{minipage}
 \caption{(a): Mean squared error for the biology data. The horizontal axis denotes the number of selected features, and the vertical axis denotes the mean squared error. Here, we use the kernel parameter $\sigma_{\mathrm x} = 1.0$ for both HSIC Lasso and NOCCO Lasso. (b),(c): Mean squared error of HSIC Lasso and NOCCO Lasso with respect to different kernel parameters. (d): Mean correlation for the biology data. The horizontal axis denotes the number of selected features, and the vertical axis denotes the mean correlation. (e),(f): Mean correlation coefficient of HSIC Lasso and NOCCO Lasso with respect to different kernel parameters. The average redundancy rate of HSIC Lasso, NOCCO Lasso, mRMR, and Lasso are 0.44, 0.45, 0.42, and 0.43. }
    \label{fig:result_TRIM32}
\end{center}
\vspace{-0.2in}
\end{figure*}

\vspace{0.1in} \noindent {\bf Results:}
Figure~\ref{fig:result_TRIM32} shows the mean squared error and the mean correlation coefficient over 100 runs as functions of the number of selected features. As can be observed, the proposed HSIC Lasso and NOCCO Lasso compare favorably with existing methods.

\section{Conclusion}
\label{sec:conclusion}
In this paper, we proposed novel non-linear feature selection methods called HSIC Lasso and NOCCO Lasso. 
In the proposed methods, global optimal solutions can be obtained
by solving a Lasso optimization problem with a non-negativity constraint,
which can be efficiently performed by 
the dual augmented Lagrangian algorithm \citep{JMLR:Tomioka+etal:2011}.
Furthermore, the proposed methods have clear statistical interpretation that
non-redundant features with strong statistical dependence on output values can be found
via kernel-based independence measures
\citep{ALT:Gretton+etal:2005,NIPS:Fukumizu+etal:2008}.
We applied the proposed methods to real-world image and biological feature selection tasks,
and experimentally showed that they are promising.

The usefulness of the proposed method will be further investigated on more real-world applications such as computer vision, bioinformatics, and speech and signal processing in the future work. Moreover, extending the proposed model to multi-task learning and prediction 
and investigating theoretical properties of the proposed formulation
are important issues to be investigated.

\section*{Acknowledgments}
The authors thank Prof.~Pradeep Ravikumar for providing us the SpAM code and Dr.~Junming Yin and Dr.~Kenji Fukumizu for their valuable comments. MY acknowledges the JST PRESTO Program and the PLIP Program, WJ acknowledges the Okazaki Kaheita International Scholarship and MEXT KAKENHI 23120004,
and MS acknowledges the FIRST program.

\bibliography{main}
\bibliographystyle{natbib}

\end{document}